\documentclass{article}
\usepackage{ijcai25}

\usepackage{times}
\usepackage{soul}
\usepackage{url}
\usepackage[colorlinks=true,linkcolor=blue,urlcolor=magenta,citecolor=darkblue]{hyperref}
\usepackage{xcolor}
\definecolor{darkblue}{rgb}{0.0, 0.0, 0.55} 
\usepackage[utf8]{inputenc}
\usepackage[small]{caption}
\usepackage{graphicx}
\usepackage{amsmath}
\usepackage{amsthm}
\usepackage{amssymb}
\usepackage{booktabs}
\usepackage{algorithm}
\usepackage{algorithmic}
\usepackage[switch]{lineno}

\usepackage{subcaption}
\usepackage{multirow}
\usepackage{multicol}
\usepackage{comment}
\usepackage[numbers]{natbib}
\usepackage{fancyhdr}
\usepackage{appendix} 
\usepackage{etoolbox}
\usepackage{cleveref} 

\patchcmd{\chapter}{\thispagestyle{plain}}{\thispagestyle{{fancy}}}{}{}

\newtheorem{remark}{Remark}
\DeclareMathOperator*{\argmin}{arg\,min}

\title{HyReaL: Clustering Attributed Graph via Hyper-Complex Space \\Representation Learning}

\author{
Junyang Chen$^1$
\and
Yang Lu$^2$\and
Mengke Li$^{3}$\and
Cuie Yang$^4$\and 
Yiqun Zhang$^5$\and
Yiu-ming Cheung$^5$\\
\affiliations
$^1$Tsinghua University
$^2$Xiamen University
$^3$Shenzhen University\\
$^4$Northeastern University
$^5$Hong Kong Baptist University
\emails
jychen152@foxmail.com, luyang@xmu.edu.cn, csmengkeli@gmail.com,
yangcuie@mail.neu.edu.cn, yqzhang@comp.hkbu.edu.hk, ymc@comp.hkbu.edu.hk
}

\begin{document}


\maketitle
\begin{abstract}
Clustering complex data in the form of attributed graphs has attracted increasing attention, where powerful graph representation is a critical prerequisite. However, the well-known Over-Smoothing (OS) effect makes Graph Convolutional Networks tend to homogenize the representation of graph nodes, while the existing OS solutions focus on alleviating the homogeneity of nodes' embeddings from the aspect of graph topology information, which is inconsistent with the attributed graph clustering objective. Therefore, we introduce hyper-complex space with powerful quaternion feature transformation to enhance the representation learning of the attributes. A generalized \textbf{Hy}per-complex space \textbf{Re}present\textbf{a}tion \textbf{L}earning (\textbf{HyReaL}) model is designed to: 1) bridge arbitrary dimensional attributes to the well-developed quaternion algebra with four parts, and 2) connect the learned representations to more generalized clustering objective without being restricted to a given number of clusters $k$. The novel introduction of quaternion benefits attributed graph clustering from two aspects: 1) enhanced attribute coupling learning capability allows complex attribute information to be sufficiently exploited in clustering, and 2) stronger learning capability makes it unnecessary to stack too many graph convolution layers, naturally alleviating the OS problem. It turns out that the node representations learned by HyReaL are more discriminative and widely suit downstream clustering with different $k$s. Extensive experiments including significance tests, ablation studies, qualitative results, etc., show the superiority of HyReaL. 
\end{abstract}

\section{Introduction}

\begin{figure}
    \centering
    \includegraphics[width=1.01\linewidth] {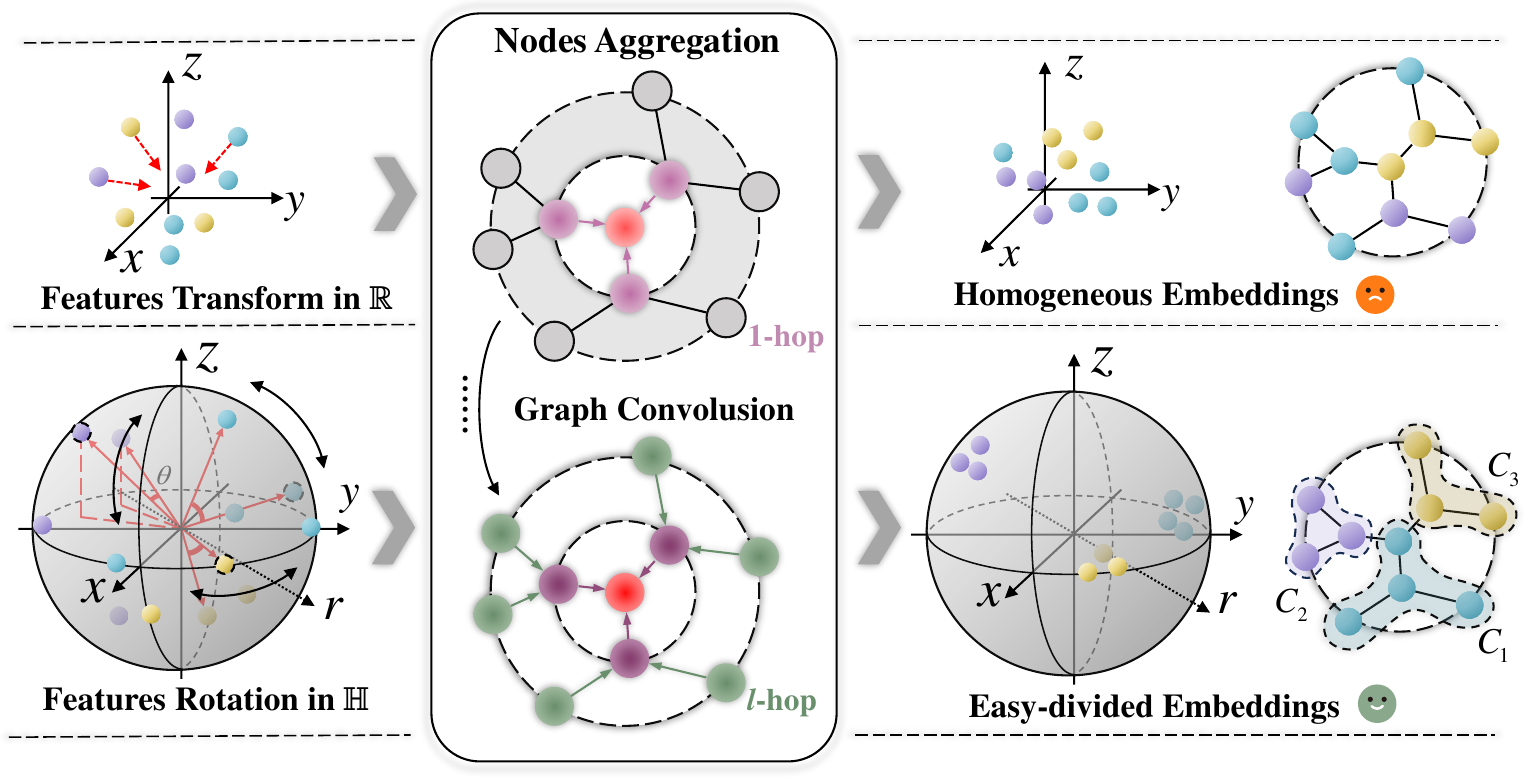}
    \caption{Vanilla graph encoders (upper) vs. quaternion graph encoders (lower). After the node information aggregation through several hops, nodes represented in real-value space $\mathbb{R}$ by vanilla graph encoders tend to be homogeneous due to the ``over-smoothing'' (OS) and ``over-dominating'' (OD) effects. By contrast, the four views of data are flexibly rotated in a hyper-complex space $\mathbb{H}$ by the quaternion graph encoders to facilitate representation learning with a higher degree of learning freedom.}
    \label{fig:intro}
\end{figure}
\label{sec-introduction}
Learning clustered distributions of data in an unsupervised setting is a fundamental data analytic process in artificial intelligence tasks. As graph data contain richer relational information among data objects (also called nodes interchangeably), clustering complex data represented in the form of graphs has attracted increasing attention \cite{deep_graph_clustering_survey,EGAE,ARGAE/ARVGAE,CCGC,mm/convert}, where graph representation \cite{shi2000normalized/SC,ng2001spectralClustering,von2007tutorial} is critical to clustering accuracy. Some recent works \cite{ARGAE/ARVGAE,deep_graph_clustering_survey} further consider attribute values of graph nodes that reflect their inherent similarity relationship to achieve a more information-comprehensive clustering. 

To perform attributed graph clustering, conventional representation learning approaches \cite{ren2020consensus,ren2021multiple} usually adopt multiple kernel functions for node embedding. However, this type of approach involves the non-trivial selection of kernels. Under such circumstances, end-to-end deep graph representation learning based on Graph Convolution Network (GCN) \cite{GCN} provides an effective way to enhance the performance of attributed graph clustering. GCN and its variants \cite{VAE,DAEGC,EGAE,R-GAE} can simultaneously learn the embedding of graph structure and attribute values to achieve a more informative representation. 

To explore the cluster distribution of data from a global perspective, relationships among nodes that span far in the graph are also critical. Although stacking more graph convolutional layers may theoretically help extract long-span information, the high overlap of over-hopped nodes will tend to homogenize the node representations, which is known as the widely discussed \emph{Over-Smoothing} (OS) effect \cite{propose/over-smooth}. Existing solutions for mitigating such effect can be categorized into training tricks \cite{regularization/EGNNs,regularization/DropEdge,regularization/PairNorm}, dynamic hopping strategies \cite{dynamic/GraphCON,dynamic/PDE-GCN}, residual connections \cite{res-con/GCNII,res-con/jknets}, and more powerful representation enhancement paradigms, e.g., contrastive learning \cite{CCGC,mm/convert}.

Most existing OS solutions focus more on relieving the smoothing of nodes in terms of topological information for serving supervised learning tasks. However, from the perspective of unsupervised clustering, overemphasizing the graph topology and the distinction of nodes may hamper the learning of compact clusters. That is, the similarity of nodes' embeddings can be easily \emph{dominated} by the topology while attribute information tends to be overlooked as the graph convolution operation is guided by the topology. As a result, such an \emph{Over-Dominating} (OD) effect will improperly drive the learning process away from the objective of clustering. Although a wide and shallow network can emphasize the coupling of attributes without incurring OS, ``shallow'' restricts the acquisition of abstract-level representations, and ``wide'' limits the scalability and efficiency of the model. In summary, conventional OS and OD solutions are contradictory, and may also introduce side effects to representation learning. Therefore, simultaneously coping with the OS and OD effects is the key to obtaining clustering-friendly representations on attributed graphs.


This paper therefore performs graph convolution in a four-axis hyper-complex space to achieve representation learning with adequate attribute interaction (counter OD) and shallow model architecture (counter OS). More detailed justifications for adopting hyper-complex space are provided in the following. Feature rotation in the hyper-complex space $\mathbb{H}$ is usually described by the Hamilton product of two quaternions~\cite{parcollet2020survey}. Figure~\ref{fig:intro} illustrates that the feature quaternion can be transformed (through the parameter quaternion of a model) to simulate feature interaction. Since the intractable Gimbal lock problem in real-valued space $\mathbb{R}$ is circumvented by the quaternion transformation, feature interaction with four times the Degrees of Freedom (DoF) under the same parameter scale can thus be achieved. This can efficiently facilitate coupling learning of input features without incurring too many extra parameters. The high DoF of quaternion transformation enhances the attributes' contribution to the representation learning, thereby considerably mitigating the OD effect, whilst the strong fitting capability brought by the quaternion operation can relieve the need to stack more GCN layers, naturally circumventing the OS issue.



As for the model process design, the framework of a graph auto-encoder \cite{kipf2016variational} is adopted to perform 
\underline{\textbf{Hy}}per-complex space \underline{\textbf{Re}}present\underline{\textbf{a}}tion \underline{\textbf{L}}earning (\underline{\textbf{HyReaL}}). We project input attributes into the hyper-complex space $\mathbb{H}$ with four views corresponding to the four parts of quaternion, in order to facilitate the efficient feature quaternion transformation through the learnable parameter quaternion of the model. Such a four-view feature amplification also helps enhance the attributes' contribution for relieving the OD effect. Through learning the parameters of the Four-View Projection (FVP) and the Quaternion Graph Encoders (QGE) with a loss composed of graph reconstruction loss and a clustering objective, the proposed HyReaL model can output clustering-friendly embeddings to fit different clustering tasks. Considering that an optimal $k$ is usually unavailable in real clustering scenarios,
we integrate a general clustering objective to the loss without requiring a pre-specified $k$. This allows the model to train more general representations and, more importantly, makes the model feasible in most cases where the `true' $k$ is unknown in advance. Theoretical analysis and extensive experiments on various real benchmark graph datasets have illustrated the superiority of HyReaL. The main contributions are summarized in four-fold:
\begin{itemize}
\item A generalized representation learning framework is proposed for attributed graph clustering. To our knowledge, this is the first attempt to: 1) introduce quaternion to unsupervised learning by bridging arbitrary dimensional attributes to the four-part quaternion algebra, and 2) connect the representation learning to clustering without requiring $k$ as a prior knowledge.
\item The OD effect is first discovered and explored in the context of graph clustering. This effect reasonably explains that the performance bottleneck encountered by current deep graph clustering is due to the disregard of the attribute information, and also provides potential guidance to subsequent graph clustering research.
\item Quaternion is novelly introduced to appropriately address both the OS and OD effects. For OD, FVP feature amplification and QGE feature coupling learning enhancement collaboratively emphasize the graph attributes. For OS, the `4$\times$DoF' of quaternion transformation ensures the learning ability of the model, making it unnecessary to tack too many GCN layers. 
\item The proposed HyReaL excels in clustering accuracy while being flexible in real applications compared to the advanced arts. That is, HyReaL learns general representations suitable for different $k$s. This avoids repeated model training when users try different $k$s to understand clusters at different granularities.
\end{itemize}

\section{Related Work}
\label{sec-related-work}

\subsection{Deep Attributed Graph Clustering}

Deep attributed graph clustering that partitions connected nodes described by attribute values into compact clusters has attracted much attention in recent years. Benefiting from the powerful representation reconstruction ability of Auto-Encoder (AE) \cite{AE} and Variational Auto-Encoder (VAE) \cite{VAE}, GAE and VGAE \cite{kipf2016variational} are proposed with graph convolution operator for graph reconstruction. Inspired by the success of GAE, recent works further improve it by introducing the attention mechanism \cite{DAEGC} and the adversarial learning mechanism \cite{ARGAE/ARVGAE}. 
To perform more accurate graph clustering, some recent works like EGAE and R-GAE \cite{EGAE}\cite{R-GAE} propose to customize the representations to be clustering-friendly by optimizing both reconstruction and clustering objectives during the model training. Most recently, contrastive learning \cite{CCGC}, as a powerful learning capability enhancement paradigm, has also been introduced to graph clustering. It adopts clustering as a proxy task for data augmentation, and generates more discriminative node embeddings. Later, a learnable reversible perturb-recover proxy task is further considered in contrastive graph clustering \cite{mm/convert}. It more reliably preserves semantic information in the augmentation, and thus achieves more satisfactory clustering performance.

\subsection{Quaternion Representation Learning}

Quaternion is a four-dimensional extension of complex numbers with a completed algebra foundation.
Since the Hamilton product \cite{quaternion-Hamilton-product} efficiently facilitates the interaction between the four parts of quaternions through the quaternion vector rotation upon the three imaginary axes, the quaternion operator is considered promising to enhance the representation learning ability \cite{parcollet2020survey}, especially for the data with natural relations among its feature tuples, e.g., the three channels of colored image \cite{eccv/QCNN,QCL/ZhengHYPLL23,QCNN4imageconstruction} and the 3D sound signals \cite{QCNN4signal}. The work \cite{nisp2019-quaternion-KG} is considered the first attempt to introduce the powerful quaternion for knowledge graph embedding. 
Almost all the existing usage of quaternion is in supervised scenarios. Moreover, the input data are usually with inherent tuple or multiple feature components corresponding to the three imaginary parts of the quaternion, and the feature components are also interdependent, especially suitable for representation learning using quaternion rotation operations. However, how to inherit the merits of quaternion to unsupervised learning tasks has been relatively unexplored.

\section{Proposed Method} \label{sec:method}

\begin{figure}[t]
\centering{\includegraphics[width=1\linewidth]{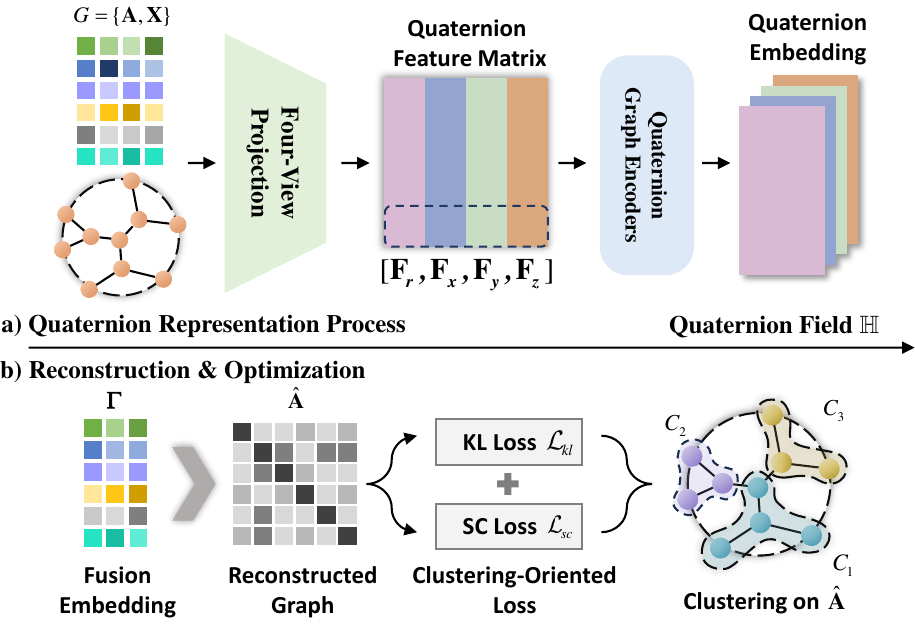}}
\caption{Overview of HyReaL. Given attributed graph $G=\{\mathbf{A},\mathbf{X}\}$, the attributes $\mathbf{X}$ are first projected into four views to form a feature quaternion $\mathbf{F}=\mathbf{F}_r+\mathbf{F}_x\mathbf{i}+\mathbf{F}_y\mathbf{j}+\mathbf{F}_z\mathbf{k}$. Then $\mathbf{F}$ is encoded with the local graph structure by a quaternion graph convolutional module. The graph clustering-friendly embedding $\boldsymbol{\Gamma}$ learned according to the joint graph reconstruction Kullback-Leibler (KL) loss $\mathcal{L}_{kl}$ and the graph clustering loss $\mathcal{L}_{sc}$ is finally obtained, which is utilized for clustering.}
\label{overall}
\end{figure}

We first provide the preliminaries, and then introduce the proposed graph clustering method HyReaL. The overview of HyReaL is demonstrated in Figure~\ref{overall}. 

\subsection{Preliminaries}

A quaternion is denoted as $ Q=r+x\boldsymbol{\rm{i}}+y\boldsymbol{\rm j}+z\boldsymbol{\rm k}$ where $r$ is the real part and $x, y, z$ are the imaginary parts. The Hamilton product between two quaternions $Q_1=r_1+x_1\boldsymbol{\rm{i}}+y_1\boldsymbol{\rm j}+z_1\boldsymbol{\rm k}$ and $Q_2=r_2+x_2\boldsymbol{\rm{i}}+y_2\boldsymbol{\rm j}+z_2\boldsymbol{\rm k}$ can be denoted as $Q_1\otimes Q_2$, which follows the laws of association and distribution, but does not follow the law of commutativity. Benefiting from the orthogonality among imaginary axes, the essence of the product is to rotate $Q_1$ according to $Q_2$ in the hyper-complex space $\mathbb{H}$ spanned by the three imaginary axes.
This is considered a critical characteristic expected by representation learning, especially for the learning of naturally coupled feature components. 

Given an undirected attributed graph $G=\{\mathbf{A},\mathbf{X}\}$ with node attributes $\mathbf{X}\in \mathbb{R}^{n\times d}$ and adjacency matrix $\mathbf{A}\in \mathbb{R}^{n\times n}$, where $n$ and $d$ are the number of nodes and dimensions, respectively. The attribute matrix can also be denoted into the form of $n$ nodes $\mathbf{X}=[\mathbf{x}_1,\mathbf{x}_2,\cdots,\mathbf{x}_n]^{\top}$, which will be grouped into $k$ clusters by partitioning the graph $\mathbf{A}$ into $k$ non-overlapping sub-graphs $\{G_1, G_2, ..., G_k\}$. A degree matrix $\mathbf{D}\in \mathbb{R}^{n\times n}$ is a diagonal matrix reflecting the connectivity of each node, which is formed by $\mathbf{D}_{ii} = \sum^{n}_{j=1} \mathbf{A}_{ij}$. The symmetric normalized Laplacian matrix of $\mathbf{A}$ that actually participates in the representation learning is denoted as $\mathbf{\tilde{A}=D^{-\frac{1}{2}} \tilde{L} D^{-\frac{1}{2}}}$, where $\mathbf{\tilde{L}}=\mathbf{I}+\mathbf{A}$ is a self-loop adjacency matrix and $\mathbf{I}$ is a unit matrix. The self-loop and normalization operations are to prevent nodes from ignoring their own information and the nodes with higher degrees from dominating the information passing during the graph convolution.

\subsection{Generalized Quaternion Representation Learning}
\label{section-HyReaL}

\paragraph{FVP: Four-View Projection}
Unlike most existing hyper-complex space representation learning scenarios that the datasets are with tuple feature components (e.g., RGB images), attributed graph data are with different numbers of attributes and various graph structures. To leverage quaternion in attributed graph representation learning, we design a learnable projection mechanism to project the attributes $\mathbf{X}$ into four views. Such a mechanism acts to lift the tuple restriction of input features in quaternion and also leverages the Hamilton product for efficient coupling learning of the attributes.

Specifically, four independent initial MLPs are utilized to project the nodes represented by the attributes $\mathbf{X}=[\mathbf{x}_1,\mathbf{x}_2,\cdots,\mathbf{x}_n]^{\top}$ into four views $\mathbf{F}_r$, $\mathbf{F}_x$, $\mathbf{F}_y$, and $\mathbf{F}_z$, which can be written as
\begin{equation}
\mathbf{F}_\triangleright=\mathcal{F}_\triangleright(\mathbf X)=\mathbf{W}^{L}_{\triangleright}\mathbf X +\mathbf{B}^{L}_\triangleright,\ \ \triangleright\in \{r,x,y,z\},
\label{multi}
\end{equation}
where $\mathcal{F}_\triangleright(\cdot)$ indicates an MLP opterator, $\mathbf{W}^{L}_{\triangleright}$ and $\mathbf{B}^{L}_\triangleright$ are the learnable parameters of an MLP. The generated four views form the feature quaternion $\mathbf F\in \mathbb{H}^{n\times(4\times \hat{d})}$ as 
\begin{equation}
\mathbf F=\mathbf{F}_r+\mathbf{F}_x\mathbf{i}+\mathbf{F}_y\mathbf{j}+\mathbf{F}_z\mathbf{k},
\label{cat}
\end{equation}
where $\hat{d}$ is the dimensionality of the features encoded by $\mathcal{F}_\triangleright(\cdot)$, and each row of $\mathbf F$ is actually the quaternion representation of the corresponding node. By introducing a learnable weight quaternion $\mathbf{W}^{Q}\in \mathbb{H}^{4d\times4\hat{d}}$ with the same size as $\mathbf{F}$, $\mathbf{F}$ can be projected based on $\mathbf{W}^{Q}$ by

\begin{small} 
\begin{equation}
\mathbf{F} \otimes \mathbf{W}^{Q} =
\left[
\begin{matrix}
\mathbf{F}_r\\
\mathbf{F}_x\\
\mathbf{F}_y\\
\mathbf{F}_z
\end{matrix}
\right ]^{\top}
\left[
 \begin{matrix}  
\mathbf{W}^{Q}_r & \mathbf {W}^{Q}_x & \mathbf {W}^{Q}_y &\mathbf {W}^{Q}_z \\
\mathbf{-W}^{Q}_x & \mathbf{W}^{Q}_r & \mathbf {-W}^{Q}_z &\mathbf{W}^{Q}_y \\
\mathbf{-W}^{Q}_y & \mathbf{W}^{Q}_z & \mathbf{W}^{Q}_r &\mathbf {-W}^{Q}_x \\
\mathbf{-W}^{Q}_z & \mathbf {-W}^{Q}_y & \mathbf{W}^{Q}_x &\mathbf{W}^{Q}_r
\end{matrix}
\right],
\label{quaternion-parameter}
\end{equation}
\end{small}

where $\otimes$ indicates the Hamilton product. 

The above $\mathcal{F}_\triangleright(\cdot)$ and $\otimes$ processes can convert an arbitrary-dimensional attribute set into the quaternion field, and associate different parts of the feature quaternion through their shared weights in the weight quaternion $\mathbf{W}^{Q}$. By tuning the weights in $\mathbf{W}^{Q}$, features in $\mathbf{F}$ can be efficiently transformed with a higher DoF to capture feature couplings. 

\begin{remark}
\textbf{Quaternion transformation facilitates a Higher Degree of Freedom (DoF).}
According to Eq.~(\ref{quaternion-parameter}), learnable parameters of $\mathbf{W}^{Q}$, i.e., $\{\mathbf{W}^{Q}_r,\mathbf{W}^{Q}_x,\mathbf{W}^{Q}_y,\mathbf{W}^{Q}_z\}$, yield 16 pairs of feature transformation to determine arbitrary rotation of $\mathbf{F}$ in hyper-complex space $\mathbb{H}$, while in real-value space $\mathbb{R}$, realizing the same transformation requires four times of parameters. Therefore, the DoF of quaternion feature transformation is four times the transformation in real-value space (detailed proof is provided in Appendix D).
\label{dof}
\end{remark}

\begin{remark}
\textbf{Benefits of high DoF in coping with OD and OS effects.}
A higher DoF can improve the learning efficiency of the model, i.e., smaller parameter scale can leverage stronger learning capability, which offers two key benefits: 1) the model can tolerate a wider structure for FVP to amplify the input attributes and thus offset the OD effect, and 2) ensuring learning capability with a wider structure can avoid OS effect caused by staking too many encoding layers.
\label{relation}
\end{remark}


\paragraph{QGE: Quaternion Graph Encoders}
To further make a fusion of the feature quaternion $\mathbf{F}$ with the graph topology $\tilde{\mathbf{A}}$, $\mathbf{F}$ is feed forward to a quaternion graph convolutional module composed of stacked encoders, where the operation of the $l$-th encoder can be written as
\begin{equation}
H_l=\varphi_l(\mathbf{\tilde A} \cdot H_{l-1} \otimes \mathbf{W}^{Q}_l)
\label{encoder}
\end{equation}
where the operation priority of $\otimes$ is higher than the matrix product.  
Here, $\varphi_l(\cdot)$ is the activation function and $\mathbf{\tilde A}\in \mathbb{R}^{n\times n}$ is the Laplacian adjacency matrix. Each encoder aggregates the $l$-hop quaternion representation of nodes according to the graph topology $\mathbf{\Tilde{A}}$, to yield a more abstract-level representation $H_l$. The output embeddings of the quaternion graph convolutional module with $m$ encoders are integrated into a single matrix $\boldsymbol \Gamma$ by 
\begin{equation}
     \boldsymbol \Gamma = {\rm Re}(H_m) \circledast {\rm Im}(H_m),
     \label{avg}
\end{equation}
where ${\rm Re}(\cdot)$ and ${\rm Im}(\cdot)$ indicate the fetch of real part and imaginary parts of $\mathbf F$, respectively, and the operation $\circledast$ is a quaternion fusion operator that takes an average of the four feature quaternion parts. Finally, we reconstruct the graph as $\mathbf {\hat A}\in \mathbb{R}^{n\times n}$ based on the embeddings $\mathbf{\Gamma}$ by
\begin{equation}
    \mathbf {\hat A} =  \boldsymbol \Gamma\cdot  \boldsymbol \Gamma^{\top}.
    \label{re}
\end{equation}
The graph reconstruction acts as a decoder to ensure the preservation of the graph topology.

\subsection{Clustering-Oriented Loss and Optimization}

From a macro perspective of the model, the FVP and QGE modules collaboratively emphasize the attribute information in $\mathbf{\Gamma}$, and the graph reconstruction acts to adapt $\mathbf{\Gamma}$ to the graph structure to seek balanced attribute and graph consensus. To also make the reconstructed graph sparse to be graph clustering friendly, the joint loss function is designed as a combination of the graph reconstruction term $\mathcal{L}_{kl}$, spectral clustering term $\mathcal{L}_{sc}$, and regularization term $\mathcal{L}_{reg}$, which can be written as
\begin{equation}
    \mathcal{L} = \mathcal{L}_{kl} + \alpha\mathcal{L}_{reg} + \beta\mathcal{L}_{sc},
    \label{loss}
\end{equation}
where $\alpha$ and $\beta$ are trade-off hyper-parameters. We adopt $\mathcal{L}_{kl}$ to quantify the reconstruction loss by
\begin{equation}
    \mathcal{L}_{kl} = \frac{1}{n^2}\sum\limits_{i=1}^{n}\sum\limits_{j=1}^{n} \mathbf{\tilde{A}}_{ij}\log \frac{1}{\mathbf{\hat{A}}_{ij}},
\end{equation}
where $n$ is the number of nodes, $\mathbf{\tilde{A}}$ and $\mathbf{\hat A}$ are the original Laplacian adjacency matrix and the adjacency matrix reconstructed by Eq.~(\ref{re}), respectively. By minimizing $\mathcal{L}_{kl}$, consensus embeddings $\mathbf{\Gamma}$ can be achieved on the graph topology reflected by $\mathbf{\tilde{A}}$ and the learned attributed graph representation indicated by $\mathbf{\hat{A}}$. The regularization term $\mathcal{L}_{reg}$ is to avoid the over-fitting of the model by restricting its complexity. The spectral clustering loss term is defined as
\begin{equation}
   \mathcal{L}_{sc} = \text{Tr} (\mathbf{\Gamma}^{\top} {\mathbf L} \mathbf{\Gamma}),
   \label{loss-sc}
\end{equation}
where $\mathbf{L} = \mathbf{D} - \mathbf{\hat{A}}$ is the Laplacian matrix formed based on the degree matrix $\mathbf{D}$ of the original graph structure $\mathbf{A}$ and the learned attributed graph representation $\mathbf{\hat{A}}$. Referring to the spectral clustering objective in Eq.~(\ref{sc-process}), $\mathbf{\Gamma}$ of $\mathcal{L}_{sc}$ can be viewed as a relaxed node-cluster affiliation indicator matrix. The minimization of $\mathcal{L}_{sc}$ prefers a $\mathbf{\Gamma}$ that can reconstruct graph with sparser adjacency (i.e., smaller values in $\mathbf{D}$) and higher feature similarity of connected nodes (i.e., larger values in $\mathbf{\hat{A}}$), both are consistent with the spectral clustering objective. 

When completing the model training, we obtain the clustering-friendly graph representation $\mathbf{\hat{A}}$ by Eq.~(\ref{re}), and the corresponding semi-definite Laplacian matrix $\mathbf{L} = \mathbf{D} - \mathbf{\hat{A}}$ is prepared to well-support the optimization of spectral clustering objective:
\begin{equation}
\argmin_{\mathbf{H}}{\rm Tr}(\mathbf{H}^{\top} \mathbf{L} \mathbf{H})\ \ \ s.t.\ \ \mathbf{H}^{\top} \mathbf{H}=\mathbf{I}.
\label{sc-process}
\end{equation}
Here $\mathbf{H}\in\mathbb{R}^{n\times k}$ is the indicator matrix indicating the node-cluster affiliations, and $\mathbf{I}$ is the unit matrix. Spectral clustering solves the above problem by first computing $\mathbf{E}$, which is the $k$-smallest eigenvectors of $\mathbf{L}$. Then K-Means clustering is performed on the $n\times k$ matrix $\mathbf{E}$ by treating each of its rows as the representation of the corresponding node. Note that the user only need to specify the target number of clusters $k$ at this time. Please refer to \cite{von2007tutorial} and \cite{kmeans-intro} for more eigenvalue decomposition and K-Means clustering details.

The whole HyReaL algorithm, its complexity analysis, and the remark that discusses the rationality of its loss function remark are provided in Appendix B.

\section{Experiment}
\label{sec-experiments}

\begin{table*}[!t]
\centering
\caption{Clustering performance compared with existing methods. The best and second-best results on each dataset are marked in \textbf{boldface} and \underline{underline}, respectively. `N/A' indicates `not available' due to gradient explosion. The ``AR'' row reports the average performance ranks.}
\resizebox{0.95\linewidth}{!}{
\begin{tabular}{c|c|ccccccccccc|c}
\toprule
Dataset & Metric & K-Means & Spectral-C & GAE & VGAE & ARGAE & ARVGAE & CONVERT & CCGC & DFCN & DAEGC & EGAE & HyReaL (ours) \\
\midrule
\multirow{3}{*}{ACM}
        & ACC & 36.78±0.01 & 74.21±0.00 & 44.22±4.11 & 59.88±1.57 & 78.56±5.10 & 86.94±1.37 & 80.53±2.91 & \underline{89.26±0.60} & 86.04±2.18 & 74.61±10.00 & 85.54±3.62 & \textbf{90.53±0.55} \\
        & NMI & 00.82±0.01 & 52.45±0.01 & 14.67±4.53 & 18.78±1.13 & 44.88±7.13 & 58.20±3.29 & 47.45±4.35 & \underline{65.36±1.21} & 59.66±4.51 & 47.92±10.35 & 56.09±8.26 & \textbf{67.67±1.24} \\
        & ARI & 00.24±0.01 & 47.65±0.00 & 03.66±2.39 & 15.59±1.86 & 46.36±11.01 & 64.94±3.29 & 51.30±6.03 & \underline{71.06±1.37} & 63.94±4.79 & 48.70±12.59 & 62.10±8.21 & \textbf{73.93±1.35} \\
\midrule
\multirow{3}{*}{WIKI}
        & ACC & 25.81±0.89 & 17.46±0.35 & 33.11±2.08 & 31.73±0.75 & 28.11±1.47 & 44.47±3.66 & \underline{51.41±1.15} & 51.29±0.84 & 43.10±3.67 & 25.38±3.35 & 47.49±1.13 & \textbf{52.59±0.88} \\
        & NMI & 22.69±1.21 & 08.84±0.16 & 31.62±1.51 & 27.25±0.38 & 23.15±1.94 & 44.13±2.65 & \underline{48.46±0.62} & 46.19±1.01 & 38.33±2.91 & 15.15±2.63 & 43.33±1.99 & \textbf{50.01±1.31} \\
        & ARI & 02.54±0.32 & -00.30±0.09 & 05.61±0.89 & 15.63±0.79 & 06.23±1.13 & 24.44±3.24 & 28.39±1.33 & 25.50±2.72 & 17.17±3.75 & 07.68±2.25 & \underline{28.99±1.58} & \textbf{34.35±1.18} \\
\midrule
\multirow{3}{*}{CITESEER}
        & ACC & 26.10±1.33 & 19.56±0.01 & 32.93±3.01 & 55.10±2.19 & 44.64±7.66 & 54.37±2.96 & 62.14±1.53 & \underline{66.31±2.27} & 42.37±2.05 & 42.66±4.74 & 58.71±3.68 & \textbf{67.41±0.63} \\
        & NMI & 06.92±1.36 & 00.31±0.00 & 20.11±2.63 & 27.92±0.86 & 19.07±6.89 & 27.54±2.85 & 34.68±1.78 & \underline{40.45±2.68} & 23.90±1.83 & 18.79±3.56 & 33.15±2.99 & \textbf{41.14±0.76} \\
        & ARI & 00.31±1.93 & 00.08±0.00 & 04.64±2.01 & 26.78±1.68 & 16.07±7.15 & 25.11±3.66 & 34.69±1.88 & \underline{39.12±3.36} & 19.19±2.43 & 16.81±4.38 & 31.46±4.61 & \textbf{42.17±0.67} \\
\midrule
\multirow{3}{*}{DBLP}
        & ACC & 32.74±0.06 & 29.92±1.01 & 46.10±1.43 & 47.07±2.43 & \underline{55.31±4.93} & 54.97±6.88 & 54.52±2.37 & 54.78±1.97 & 38.91±0.04 & 43.36±4.72 & 53.64±1.46 & \textbf{72.45±1.66} \\
        & NMI & 02.98±0.01 & 00.28±0.22 & 19.71±1.83 & 17.72±2.11 & 20.63±3.63 & 22.61±5.44 & 22.33±1.93 & \underline{23.81±2.53} & 08.11±0.04 & 11.41±3.55 & 18.19±1.07 & \textbf{39.58±1.91} \\
        & ARI & 15.31±1.87 & 00.20±2.80 & 05.78±0.87 & 14.39±1.95 & 18.14±4.36 & 17.70±5.12 & 17.81±1.17 & \underline{18.64±1.28} & 06.63±0.02 & 10.40±3.70 & 15.07±2.02 & \textbf{40.96±2.79} \\
\midrule
\multirow{3}{*}{FILM}
        & ACC & 24.21±0.01 & 24.05±0.04 & 25.64±0.02 & 21.40±0.79 & 23.84±0.47 & 24.31±1.32 & \textbf{27.43±0.23} & 26.36±0.11 & 25.91±1.64 & 24.61±0.33 & 22.79±0.25 & \underline{26.84±0.83} \\
        & NMI & 00.01±0.00 & 00.11±0.01 & 00.09±0.01 & 00.07±0.01 & 00.16±0.05 & 00.22±0.39 & \underline{00.79±0.07} & 00.15±0.01 & 00.28±0.03 & 00.09±0.03 & 00.21±0.08 & \textbf{1.39±0.28} \\
        & ARI & 00.00±0.01 & -00.14±0.02 & 00.13±0.01 & 00.01±0.02 & 00.11±0.03 & 00.31±0.51 & \underline{01.34±0.17} & 00.24±0.05 & 00.27±0.04 & 00.15±0.10 & 00.17±0.08 & \textbf{1.74±0.45} \\
\midrule
\multirow{3}{*}{CORNELL}
        & ACC & \textbf{42.40±0.65} & 37.81±2.34 & 38.03±1.09 & 26.66±1.16 & 36.99±2.54 & 36.55±2.65 & \underline{41.86±2.98} & 39.61±2.09 & 39.72±1.90 & 36.28±2.11 & 39.23±0.53 & 37.65±1.33 \\
        & NMI & 02.71±0.17 & 03.69±0.62 & 05.35±0.36 & 03.25±0.97 & 06.01±1.22 & 03.19±0.56 & \textbf{09.80±2.68} & 04.89±1.04 & 03.25±0.37 & 06.83±1.36 & 06.49±0.73 & \underline{8.14±1.86} \\
        & ARI & -02.14±0.10 & -00.15±0.55 & 02.11±0.48 & -00.06±0.54 & 02.43±1.96 & 00.85±1.18 & \textbf{06.19±3.25} & 02.07±1.06 & -01.10±1.23 & 02.15±2.08 & 03.17±0.91 & \underline{5.40±1.87} \\
\midrule
\multirow{3}{*}{CORA}
        & ACC & 31.14±3.76 & 24.47±0.03 & 49.47±5.76 & 63.47±0.69 & 65.96±4.12 & 66.72±3.04 & 66.34±1.80 & 72.00±1.77 & 45.94±5.80 & 45.30±5.92 & \underline{72.11±1.35} & \textbf{73.28±1.61} \\
        & NMI & 06.67±5.28 & 01.48±0.01 & 40.86±4.81 & 45.45±0.59 & 44.75±3.69 & 48.96±2.62 & 46.84±1.68 & \underline{55.02±1.91} & 36.46±3.44 & 25.88±4.35 & 52.89±1.17 & \textbf{56.22±1.21} \\
        & ARI & 07.83±1.69 & -00.08±0.01 & 22.49±7.27 & 39.01±0.85 & 39.52±4.55 & 42.80±2.69 & 40.13±1.67 & \underline{49.17±2.40} & 23.95±5.52 & 20.09±5.99 & 48.49±2.16 & \textbf{51.34±2.00} \\
\midrule
\multirow{3}{*}{WISC}
        & ACC & 42.03±2.04 & 30.31±0.11 & 42.11±1.73 & 25.77±1.34 & 36.01±2.18 & 37.17±2.58 & \textbf{47.61±1.91} & \underline{44.14±0.86} & 40.95±5.44 & 25.38±3.35 & 37.01±2.01 & \underline{44.14±1.88} \\
        & NMI & 06.25±1.13 & 03.73±0.01 & 08.09±0.63 & 02.60±1.40 & 11.02±2.61 & 05.35±3.26 & 09.70±3.35 & 08.39±0.45 & 07.01±0.58 & \underline{15.15±2.63} & 11.02±1.16 & \textbf{16.59±3.31} \\
        & ARI & -03.02±1.68 & 00.02±0.01 & 02.85±0.54 & 00.03±0.48 & 05.56±1.86 & 02.02±1.63 & 04.76±2.69 & 03.60±0.90 & 04.45±0.99 & \underline{07.68±2.25} & 06.00±1.12 & \textbf{10.38±1.76} \\
\midrule
\multirow{3}{*}{UAT}
        & ACC & 32.69±0.12 & 32.52±0.01 & 44.55±0.07 & 37.45±3.46 & 49.36±1.30 & 41.85±1.63 & \textbf{55.18±1.34} & 47.88±2.69 & 39.33±4.72 & 52.49±1.25 & 53.10±0.79 & \underline{53.82±0.54} \\
        & NMI & 20.63±0.63 & 03.43±0.00 & 18.61±9.44 & 17.68±0.95 & 23.33±1.71 & 15.86±2.38 & \textbf{27.31±1.18} & 20.63±2.64 & 13.95±1.67 & 21.42±1.29 & 21.80±0.85 & \underline{24.03±1.06} \\
        & ARI & 06.42±0.44 & 01.57±0.00 & 11.61±5.90 & 14.35±0.84 & 16.76±0.68 & 10.33±2.82 & 19.46±1.90 & 12.95±1.80 & 07.28±3.16 & \underline{21.07±1.11} & 20.77±0.64 & \textbf{22.38±0.78} \\
\midrule
\multirow{3}{*}{AMAP}
        & ACC & 22.66±0.31 & 17.24±0.01 & 60.47±0.87 & 68.61±0.51 & N/A & N/A & 66.28±1.86 & \textbf{77.07±0.38} & 58.51±3.96 & 47.45±3.36 & 76.37±1.32 & \underline{77.00±0.98} \\
        & NMI & 02.37±0.09 & 00.53±0.00 & 58.01±0.56 & 55.04±0.46 & N/A & N/A & 52.57±0.79 & \underline{67.06±0.72} & 55.95±1.13 & 38.83±4.24 & 65.44±1.61 & \textbf{69.01±1.60} \\
        & ARI & 00.37±0.03 & 00.00±0.01 & 33.31±1.02 & 46.55±0.67 & N/A & N/A & 42.89±1.49 & \underline{57.55±0.44} & 41.76±1.58 & 25.03±4.68 & 57.51±1.74 & \textbf{59.85±1.49} \\
\midrule
        - & AR & 9.9 & 10.6 & 7.8 & 8.2 & 6.0 & 5.4 & 3.6 & 3.7 & 7.0 & 7.6 & 4.4 & 1.5 \\
\bottomrule
\end{tabular}
}
\label{compare-result}
\end{table*}

\subsection{Experimental Settings}
\label{subsec-settings}

\begin{table*}[t]
\centering
\caption{Ablation study of the key modules of HyReaL. The best and second-best results are marked in \textbf{boldface} and \underline{underline}, respectively.}
\resizebox{0.9\linewidth}{!}{
\begin{tabular}{c|ccc|ccc|ccc|ccc|ccc}
\toprule
\multirow{2}{*}{\centering Dataset} & \multicolumn{3}{c|}{Baseline} & \multicolumn{3}{c|}{HyReaL w/o FVP} & \multicolumn{3}{c|}{HyReaL w/o QGE} & \multicolumn{3}{c|}{HyReaL w/o $\beta$} & \multicolumn{3}{c}{\textbf{HyReaL}} \\
\cmidrule(lr){2-4} \cmidrule(lr){5-7} \cmidrule(lr){8-10} \cmidrule(lr){11-13} \cmidrule(lr){14-16}
  & ACC & NMI & ARI & ACC & NMI & ARI & ACC & NMI & ARI & ACC & NMI & ARI & ACC & NMI & ARI \\
\midrule
        ACM & 89.34 & 64.54 & 70.92 & 89.90 & 65.80 & 72.27 & 84.53 & 56.38 & 60.85 & \underline{90.44} & \underline{67.48} & \underline{73.70} & \textbf{90.53} & \textbf{67.67} & \textbf{73.93}  \\

        WIKI & 51.60 & 49.15 & 32.43 & 51.57 & 47.91 & 31.99 & 51.64 & 48.66 & 32.45 & \textbf{53.16} & \textbf{50.13} & \underline{34.24} & \underline{52.59} & \underline{50.01} & \textbf{34.35}  \\

        CITESEER & 65.73 & 40.24 & 40.72 & 66.21 & 40.44 & 41.28 & 66.57 & 40.38 & 41.35 & \underline{67.31} & \underline{41.08} & \underline{42.07} & \textbf{67.41} & \textbf{41.14} & \textbf{42.17}  \\

        DBLP & 67.89 & 35.20 & 35.48 & \underline{71.41} & \underline{38.15} & \underline{39.73} & 67.26 & 36.59 & 35.71 & 69.93 & 37.80 & 38.15 & \textbf{72.45} & \textbf{39.58} & \textbf{40.96}  \\

        FILM & 26.95 & 1.12 & 1.76 & 27.41 & 1.28 & \underline{1.97} & \textbf{27.70} & \textbf{1.51} & \textbf{2.01} & \underline{27.42} & \underline{1.49} & 1.86 & 26.84 & 1.39 & 1.74  \\

        CORNELL & 35.85 & 6.87 & 3.69 & 36.99 & 6.52 & 4.51 & 36.61 & 6.52 & 3.93 & \textbf{38.14} & \textbf{9.37} & \underline{4.99} & \underline{37.65} & \underline{8.14} & \textbf{5.40}  \\

        CORA & 72.73 & \underline{55.84} & \textbf{51.44} & \underline{73.12} & 55.13 & 50.61 & 72.11 & 55.35 & 49.68 & 70.89 & 53.25 & 48.05 & \textbf{73.28} & \textbf{56.22} & \underline{51.34}  \\

        WISC & 40.92 & 13.54 & 8.03 & 43.03 & 16.09 & 9.37 & \textbf{44.74} & \textbf{17.21} & \textbf{10.45} & 43.63 & 16.40 & 9.78 & \underline{44.14} & \underline{16.59} & \underline{10.38}  \\

        UAT & 53.68 & 23.69 & 21.87 & 53.71 & \underline{23.77} & \underline{22.36} & \textbf{54.37} & 23.26 & 22.19 & 53.47 & 23.68 & 22.11 & \underline{53.82} & \textbf{24.03} & \textbf{22.38}  \\

        AMAP & 73.23 & 61.13 & 53.81 & 74.69 & 63.65 & 55.53 & 74.98 & 64.34 & 56.37 & \underline{77.00} & \underline{69.01} & \underline{59.85} & \textbf{77.00} & \textbf{69.01} & \textbf{59.85}  \\

\midrule
        Average Rank & \multicolumn{3}{c|}{4.20} & \multicolumn{3}{c|}{3.33} & \multicolumn{3}{c|}{3.17} & \multicolumn{3}{c|}{2.60} & \multicolumn{3}{c}{\textbf{1.67}} \\
\bottomrule
\end{tabular}
}
\label{ablation-study}
\end{table*}

\paragraph{Datasets} 
Experiments are conducted on ten real benchmark attributed graph datasets, including CORA \cite{cora_citeseer}, CITESEER \cite{cora_citeseer}, DBLP \cite{DBLP_ACM}, ACM \cite{DBLP_ACM}, WIKI \cite{wiki}, FILM \cite{deep_graph_clustering_survey}, and the four, i.e., CORNELL, WISC, UAT, and AMAP, from \cite{githubdataset/deep_graph_clustering_survey}. The CORA and DBLP datasets are the citation network. ACM and DBLP datasets are paper citation relationships. WIKI and FILM datasets are the relationships of Wikipedia links and films, respectively. CORNELL, WISC, UAT, and AMAP datasets are American university website links. Detailed dataset statistics are sorted in the Appendix A.

\paragraph{Training Process}
All the experiments are implemented in PyTorch 1.8.0 on NVIDIA A5000 GPU, 64GB RAM. We first warm up the model by a 10-epoch training using only the KL loss $\mathcal{L}_{kl}$ and regularization loss $\mathcal{L}_{reg}$. We follow the most recent graph clustering works \cite{CCGC,EGAE,DAEGC,DFCN,mm/convert} to obtain the clustering performance: Each result is the average performance with standard deviation on ten implementations of the compared methods. For each implementation, the model is trained by 50 epochs. In each epoch, we train the model in four iterations and then perform clustering. The best clustering performance of the 50 epochs is chosen to be the performance of the current implementation.

\paragraph{Counterparts Setup}
11 clustering methods are compared, including two traditional methods, i.e., K-Means \cite{kmeans/hamerly2003learning} and Spectral Clustering (Spectral-C, to distinguish from the internal evaluation metric SC) \cite{shi2000normalized/SC}, two conventional representation learning-based clustering methods, i.e., GAE \cite{kipf2016variational} and VGAE \cite{kipf2016variational}, seven state-of-the-art deep clustering methods including ARGAE and ARVGAE \cite{ARGAE/ARVGAE}, CONVERT \cite{mm/convert}, CCGC \cite{CCGC}, DFCN \cite{DFCN}, DAEGC \cite{DAEGC}, and EGAE \cite{EGAE}. We let K-Means directly perform clustering on the data attributes. All the other methods obtain node representations first and then implement K-means on the representations. 
Settings of all the compared methods and HyReaL are reported in Appendix A.

\paragraph{Validity Metrics}
Six evaluation metrics are utilized. Three external metrics \cite{metrics}: Clustering Accuracy (ACC), Normalized Mutual Information (NMI), and Average Rand Index (ARI), which evaluate performance according to the data labels, are in the intervals $[0,1]$, $[0,1]$, and $[-1,1]$, respectively. Three internal metrics: Silhouette Coefficient (SC) \cite{metric/sc}, Davies-Bouldin Index (DBI) \cite{metric/dbi}, and Calinski-Harabasz Index (CHI) \cite{metric/chi}, that do not rely on the labels, are in the intervals $[-1, 1]$, $[0, +\infty)$, and $[0, +\infty)$. All these metrics are commonly used by most of the compared state-of-the-art methods, and except for DBI, a higher value indicates a better clustering performance.

\subsection{Quantitative Results}

We conduct four groups of quantitative experiments: 1) Compare clustering performance using external metrics to illustrate the clustering accuracy superiority of HyReaL; 2) Compare clustering performance using internal metrics under different $k$s to verify the separability and universality of the embeddings learned by HyReaL; 3) Compare execution time to validate the efficiency of HyReaL; 4) Compare different ablated versions of HyReaL to prove the effectiveness of its core modules.

\paragraph{Clustering Performance Evaluated by External Metrics}

\Cref{compare-result} reports the clustering performance of all the compared methods by using $k$ provided by the data labels. The significance test described in Appendix A is also conducted, and the results shown in Appendix C demonstrate that HyReaL passes all the Wilcoxon signed rank tests with a confidence interval of 95\% (except for the CONVERT method in terms of ACC), which validates its superiority. From \Cref{compare-result}, it can be observed that the proposed HyReaL outperforms the compared methods in most cases. Out of the 294 comparisons, HyReaL won 290 times, which generally demonstrates its superiority. Note that six `N/A' cases happened when implementing ARGAE and ARVGAE on the AMAP dataset, as they suffered from gradient explosions. 
Finally, another three key observations are provided in the Appendix C.

\paragraph{Separability and Universality Evaluation of Representations}

\begin{figure}[!t]
    \raggedright
        \includegraphics[width=1\linewidth]{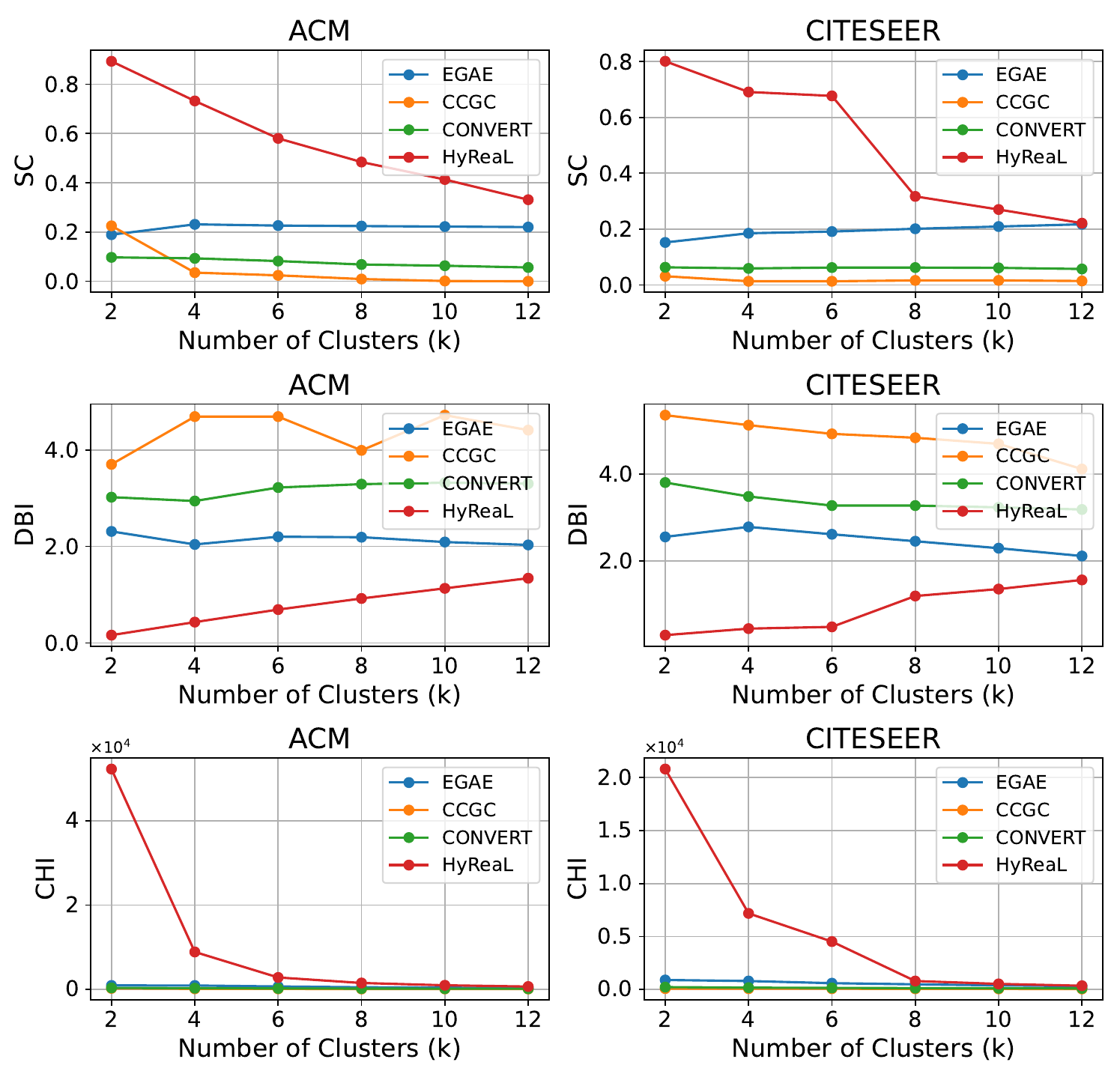}
    \caption{Clustering performance comparison using internal metrics under different $k$s. For the SC and CHI metrics, the higher the better. For the DBI metric, the lower the better.}
    \label{different-k-clustering}
\end{figure}

For clustering, the separability of learned representations is often evaluated by internal metrics. To also verify the effectiveness of the generalized loss of HyReaL, we compare the clustering performance under different $k$s of HyReaL with EGAE, CCGC, and CONVERT, which are the state-of-the-art counterparts that performed better in Table~\ref{compare-result}. The comparison results are shown in Figure~\ref{different-k-clustering}, and it can be seen that the proposed HyReaL always outperforms the state-of-the-art counterparts under different $k$s w.r.t. all the metrics. Such results simultaneously prove the outstanding separability and universality of the embeddings learned by the HyReaL in general. More specific observations are as follows: 1) As the value of $k$ increases, the performance of HyReaL gradually degrades. This is reasonable because the internal metrics mainly measure the separation between clusters and the compactness within clusters. When there are many clusters (large $k$), all the clusters tend to be smaller and compact preferred by the internal metrics, and thus the evaluation results will become similar; 
2) $k$s in Figure~\ref{different-k-clustering} cover the `true' $k$s provided by the dataset labels in Dataset Table in the Appendix A. At the `true' $k$ of each dataset, HyReaL performs better, proving the better separability of its representation.
More experiment results of different datasets are in the Appendix C. 

\paragraph{Effectiveness of QGE on Mitigating Over-smoothing}
We conduct experiments by stacking QGE layers to verify their effectiveness in mitigating the OS effect. The model architecture starts with an input dimension of 512, followed by a layer that reduces it to 256, then another layer that reduces it further to 128, which is the final output dimension. This represents the architecture of a two-layer model. Then, we stack additional QGE layers with the same input and output dimensions of 256 to build the representation sequence. 
Part of results are plotted in Figure~\ref{sub-over-smoothing-ablation} (The complete results are shown in Appendix C), where we observed a clear decreasing trend in all metrics across different datasets. These trends indicate that the deeper convolutional graph model is definitely affected by the OS effect, which leads to homogeneous representations and worse clustering performance. Furthermore, in specific datasets such as ACM and AMAP, the results become more unstable, reflecting that over-smoothing causes the randomness of the representation. In summary, all the results clearly support the conclusion that introducing high DoF quaternion representation learning alleviates the OS problem and contributes to stronger feature interaction representation.

\begin{figure}
  \centering
  \includegraphics[width=1\linewidth]{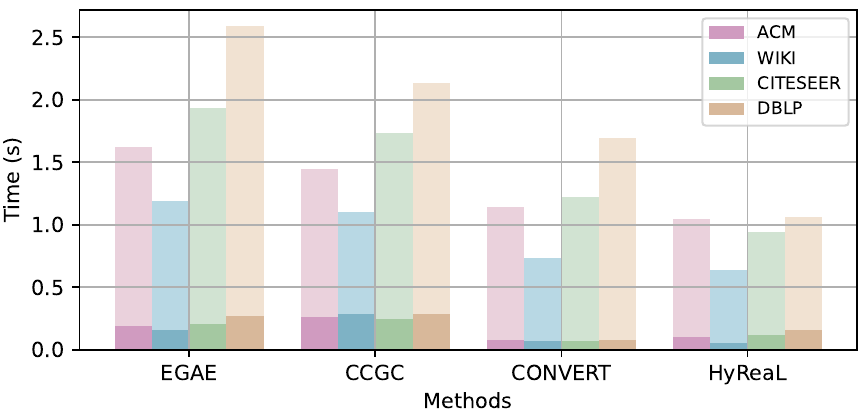} 
  \caption{Average execution time on different $k$ values. Different colors indicate the average execution time on different datasets. Deep and shallow colors indicate the execution time of model training and clustering.} 
  \label{efficiency-time} 
\end{figure}

\paragraph{Efficiency Evaluation}
Corresponding to the four datasets and three advanced methods in the previous experiment as shown in Figure~\ref{different-k-clustering} (Complete results are shown in Appendix C), we also compare their execution time with the proposed HyReaL averaged on the six different $k$ values. The execution time comparison is visualized in Figure~\ref{efficiency-time}. It can be observed that the execution time of the compared methods is higher than that of HyReaL. This is because HyReaL can learn general representations to support the clustering under different $k$s without retraining the model, and thus its model training time averaged on the six runs of the clustering is relatively lower. By contrast, the counterparts require $k$ as an input, and thus they have to re-train the model for different $k$s, incurring extra training overhead.

\paragraph{Ablation Studies}
Table~\ref{ablation-study} compares the clustering performance of HyReaL with: 1) Baseline: one MLP and two stacked GCN encoders, 2) HyReaL w/o FVP: replace the FVP module by one linear layer, 3) HyReaL w/o QGE: replace the QGE module with two conventional GCN encoders. 4) HyReaL w/o $\beta$: set the loss term $\mathcal{L}_{sc}$ to 0.
By comparing HyReaL with HyReaL w/o FVP, HyReaL w/o QGE, HyReaL w/o $\beta$, and Baseline, the effectiveness of the designed FVP and clustering-oriented loss, the necessity of introducing QGE, and the adaptability of FVP and QGE, can be validated, respectively. The effectiveness of the FVP and QGE modules can also validate the overall design in mitigating the OD effect as they serve to enhance the preservation and extraction of the attribute information.
The Appendix of complementary ablation study results provides a detailed discussion of the three key observations drawn from the ablation study comparing HyReaL with its various ablated versions.

\subsection{Qualitative Results}

\begin{figure}[!t]
    \raggedright
        \includegraphics[width=0.5\textwidth]{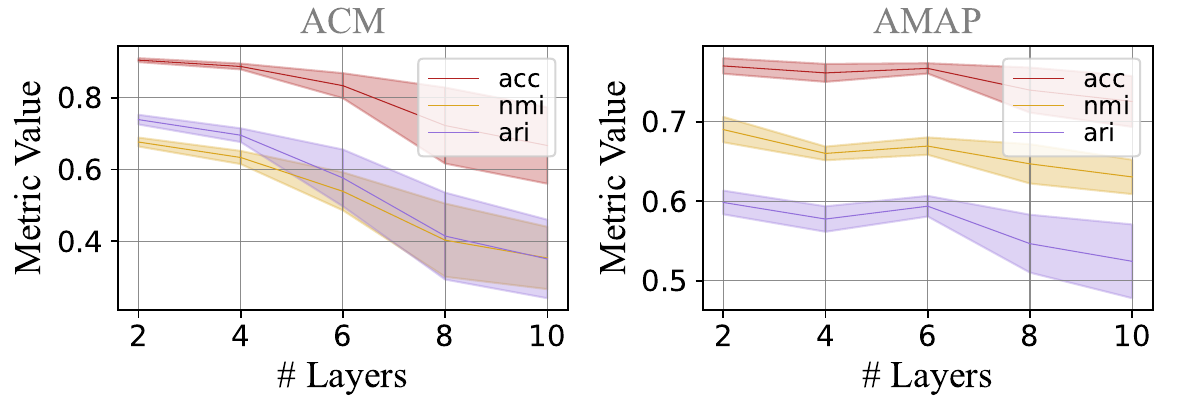}
    \caption{The OS effect analysis across ten datasets via QGE with varying layer numbers. When the number of layers increases, the OS effect significantly impacts clustering performance, highlighting the effectiveness of our method in alleviating OS.}
    \label{sub-over-smoothing-ablation}
\end{figure}

\begin{figure}[!t]
        \centering
    \includegraphics[width=1\linewidth]{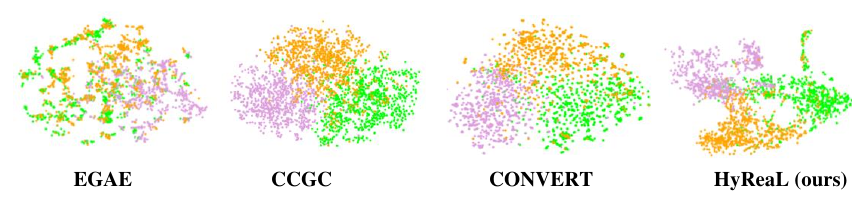}
    \caption{$t$-SNE visualization of ACM dataset represented by the four advanced methods.}
    \label{main-visual}
\end{figure}

To intuitively show the representation effectiveness of HyReaL, we visualize the distributions of embeddings generated by the state-of-the-art EGAE, CCGC, CONVERT, and our HyReaL on the ACM dataset in Figure \ref{main-visual}. The 2-D plots are generated using $t$-SNE \cite{tsne} and we use different colors to mark the label-provided clusters. Intuitively, CCGC, CONVERT, and our HyReaL perform better with more separable clusters against the EGAE. The reason would be that CCGC and CONVERT adopt contrastive augmentation, and HyReaL adopts quaternion rotation, both effectively enhance the learning capability of the corresponding representation learning models. Since HyReaL performs structural rotation of the four views of attributes, the global distribution of nodes is better preserved, and thus the embedding clusters of HyReaL are even more separable compared to that of CCGC and CONVERT. The GAE-based EGAE method is probably over-dominated by the graph topology as it does not specifically emphasize the preservation of attribute information. 

\section{Concluding Remarks}
\label{sec-conclusion-limitations}
This paper has proposed the novel HyReaL clustering method. It leverages the advantages of the efficient Hamilton product of quaternions to simultaneously tackle the OS and OD issues that bottleneck the clustering performance. Through generalized design, a representation learning model composed of learnable FVP and QGE is formed for clustering-friendly representation learning. The FVP module bridges the gap between any dimensional attributes and the four-part quaternion operation of QGE, and these two modules collaboratively enhance: 1) the learning capability of the model, and 2) the preservation of attribute information. The generalized clustering objective loss guides the model to learn universal representations with high DoF without restricting the embeddings to concentrate on a pre-specified number of clusters. As a result, HyReaL can obtain more discriminative and clustering-friendly node representations that are consistent for different $k$s. This is considered to be an important advantage for real applications and data understanding. Extensive experiments have shown the superiority of HyReaL. 
While HyReaL proves effective, it is not exempt from limitations. That is, the generality and efficiency of HyReaL are for different sought numbers of clusters $k$ on static data. Our future research will focus on improving the proposed quaternion representation learning for the adaptation of streaming data or even data with concept drift.

\bibliographystyle{named}
\bibliography{ijcai25}



\appendix
\onecolumn

\renewcommand{\thefigure}{A.\arabic{figure}}
\renewcommand{\thetable}{A.\arabic{table}}
\setcounter{figure}{0}
\setcounter{table}{0}
\setcounter{remark}{0}
\setcounter{equation}{0}

\section{Experimental Settings}
\label{app-sec-settings}
\subsection{Dataset Summary}
\begin{table}[h]
    \centering
    \caption{The statistics of ten graph datasets. $n$ is the number of nodes, $d$ is the dimension of attributes, and $k$ is the number of clusters provided by the labels of datasets.} 
    \begin{tabular}{c|c|c|c|c}
    \toprule
        No. & Dataset & $n$ & $d$ & $k$ \\ \cmidrule(lr){1-5}
        1 & ACM & 3025 & 1870 & 3 \\ 
        2 & WIKI & 2405 & 4973 & 17 \\ 
        3 & CITESEER & 3327 & 3703 & 6 \\ 
        4 & DBLP & 4057 & 334 & 4 \\ 
        5 & FILM & 7600 & 932  & 5 \\        
        6 & CORNELL & 183 & 1703 & 5 \\ 
        7 & CORA & 2708 & 1433 & 7 \\         
        8 & WISC & 251 & 1703 & 5 \\ 
        9 & UAT & 1190 & 239 & 4 \\ 
        10 & AMAP & 7650 & 745 & 8 \\
        \bottomrule
    \end{tabular}
    \label{graph-dataset}
\end{table}
The Table~\ref{graph-dataset} shows the statistical summary of used attributed graph datasets.

\subsection{Counterparts and HyReaL Settings}
\label{model-settings}
In Comparison approaches, we follow their original settings. For traditional methods, the K-Means clusters the features without graph structure, and the Spectral clustering clusters the graph structure without feature matrix. They are executed 10 times for average scores. For conventional methods, we perform 200 epochs of unsupervised training of the GAE and VGAE, then use K-Means to cluster the generated embedding. For advanced and state-of-the-art clustering approaches, we reproduce their source code by following the original parameter settings in the source codes.

There are some hyper-parameters and settings of our method, i.e., the layer number, pre-training learning rate, pre-training iteration number, learning rate, iteration number, model regularization trade-off $\alpha$, and representation embedding loss trade-off $\beta$. We set Adam optimizer during experiments. 
The activation function of the graph encoder is ReLU for each layer.
$\mathcal{L}_{reg}$ in the loss is the regularization of model, and L1 regularization is utilized. 
In the pre-training process, the hyper-parameter $\beta$ is set to $0.0001$. 
For ten datasets, the neuron number of layers, the pre-training learning rate, pre-training iteration number, and iteration number are set to $[512, 256, 128]$, $10^{-4}$, $10$, and $4$, respectively. 
The learning rate is set to $0.00001$ for CITESEER, $0.0002$ for ACM, WIKI, and AMAP, $0.0004$ for DBLP, FILM, CORNELL, CORA, WISC, and, UAT. 
The $\alpha$ is set to $10^{-4}$ for CORNELL, $2\times10^{-4}$ and UAT, $5\times10^{-4}$ for CITESEER, FILM, and WISC, $10^{-5}$ for CORA and DBLP, $10^{-6}$ for ACM, WIKI, AMAP. 
The $\beta$ is set to $2^{-10}$ for WIKI, DBLP, CORNELL, CORA, WISC, $2^{-12}$ for ACM, CITESEER, $2^{-30}$ for UAT, $0$ for AMAP.

\subsection{Description of Validity Metrics}
We provide a more detailed description of validity metrics, which are Accuracy (ACC), Normalized Mutual Information (NMI), and Adjusted Rand Index (ARI) \cite{DBLP:journals/corr/abs-2206-07579}. 

ACC is a straightforward measure that calculates the percentage of correctly classified data points in the clustering results compared to ground truth. A higher accuracy indicates better performance. Given ground truth labels $Y=\{y_i|1\le i \le n\}$ and the predicted clusters $\hat{Y}=\{\hat{y_i}|1\le i \le n\}$, ACC is computed as
\begin{equation}
    ACC(\hat{Y}, Y)=\max \frac{1}{n}\sum\limits_{i=1}^n1\{y_i=\hat{y_i}\}.
\end{equation}

NMI quantifies the amount of shared information between two clusters. It ranges from 0 to 1, where 1 indicates perfect agreement and vice versa. Higher NMI values indicate better clustering performance. The NMI can be computed by
\begin{equation}
   NMI(\tilde{Y},Y)=\frac{{T}(\tilde{Y};Y)}{\frac{1}{2}\left[H(\tilde{Y})+H(Y)\right]},
\end{equation}
where $H(Y)$ is entropy of $Y$ and $T(\tilde{Y};Y)$ is mutual information between $\tilde Y$ and $Y$.

ARI measures the similarity between two clusters, taking into account both true positive and true negative matches while correcting for chance. It produces a value between -1 and 1. An ARI value close to 1 suggests strong agreement, close to 0 indicates random agreement, and negative values indicate disagreement. A higher ARI value indicates better clustering performance, and the ARI can be computed as
\begin{equation}
   ARI = \frac{RI-\mathbb{E}(RI) }{\max (RI-\mathbb{E}(RI))},
\end{equation}
where
\begin{equation}
    RI = \frac{TP+TN}{C_n^2}.
\end{equation}
Here, $TP$ and $FP$ respectively denote the number of true positive pairs and true negative pairs, and $C_n^2$ is the number of possible object pairs. 

\subsection{Settings of the Wilcoxon Signed-ranks Test}
\label{sec-wilcoxon}

\begin{table*}[t]
\centering
        \caption{The Wilcoxon signed rank test with 95\% confidence interval. The symbols ``$+$'' and ``$-$'' indicate the rejection and acceptance of the null hypothesis.}

    \begin{tabular}{c|c|c|c}
    \toprule
        Method & ACC & NMI & ARI \\ 
        \cmidrule(lr){1-4}
        K-Means & $+$ & $+$ & $+$ \\ 
        Spectral-C  & $+$ & $+$ & $+$ \\ 
        GAE  & $+$ & $+$ & $+$ \\ 
        VGAE  & $+$ & $+$ & $+$ \\ 
        ARGAE  & $+$ & $+$ & $+$ \\ 
        ARVGAE  & $+$ & $+$ & $+$ \\ 
        CONVERT & $-$ & $+$ & $+$ \\ 
        CCGC  & $+$ & $+$ & $+$ \\ 
        DFCN  & $+$ & $+$ & $+$ \\ 
         DAEGC  & $+$ & $+$ & $+$ \\ 
        EGAE & $+$ & $+$ & $+$ \\ 
        \bottomrule
    \end{tabular}
    \label{wtest}
\end{table*}

Here, we provide experimental settings of the Wilcoxon signed-ranks test for the results in Table~\ref{wtest} of the submitted paper.

The Wilcoxon signed-ranks test is a non-parametric alternative to the paired t-test. It ranks the differences in performances of two classifiers for each dataset, ignoring the signs, and compares the ranks for the positive and the negative differences \cite{DBLP:journals/jmlr/Demsar06}. In general, the Wilcoxon signed-ranks test is used when we have paired data and try to observe if there is a significant change. If the test statistic is smaller than the critical value from a table (or if the p-value is below a chosen significance level), we can reject the null hypothesis, which suggests a significant difference between the paired data.

The procedures of the Wilcoxon signed-ranks test are: 1) Calculate the differences between paired observations. 2) Rank these differences in absolute rank values. 3) Assign positive or negative signs to the ranks based on the direction of the differences. 4) Sum the ranks of positive and negative differences separately. The smaller of the two sums is utilized for the test. If the smaller value is smaller than the critical value, we will reject the null hypothesis.

In our experiment, the Wilcoxon signed-ranks test is conducted to compare our method with other methods under different validity metrics on all the ten datasets. The procedures are as follows: 1) Formulate the hypothesis where the null hypothesis is that HyReaL does not exhibit a significant difference, or perform equally, compared to other models under a specific validity metric. The alternative hypothesis is that HyReaL significantly outperforms other models. 2) Set the significance level at 0.01. 3) Calculate the p-value of the compared model performance. 4) Obtain the test results. If the p-value is less than the chosen significance level, we reject the null hypothesis, and vice versa, where a rejection suggests that HyReaL significantly outperforms the compared model.

\section{Algorithm and Complexity Analysis of the HyReaL}
\subsection{Algorithm of HyReaL}
\label{sec-algorithm}
The algorithm process of HyReaL is shown in Algorithm~\ref{alg:algorithm}.

\subsection{Rationality of Loss Function}
\begin{remark}
    \textbf{Rationality of the loss function.}
    It is noteworthy that $\mathbf{L}$ obtained based on $\hat{\mathbf{A}}$ is the key factor to influence the accuracy of clustering. To learn more powerful $\hat{\mathbf{A}}$, the model is designed with a higher DoF in feature encoding facilitated based on the quaternion product. On such basis, the training process comprehensively takes into account the attribute information by the FVP and QGE, preserves the graph topology by the graph reconstruction decoder, and customizes the general clustering-friendly representation by introducing the clustering-oriented loss $\mathcal{L}_{sc}$.
\end{remark}

\subsection{Computational Complexity Analysis}
\label{sec-complexity}
The time complexity of the proposed HyReaL model is $\mathcal{O}(T[nd\hat{d}+n^2\hat{d}^2])$. We analyze it below. The training process of the model is composed of three parts: (1) quaternion projection, (2) quaternion graph convolution, and (3) graph reconstruction. For the quaternion projection, the dimensions of the input and projected features of each projector are $d$ and $\hat{d}$, respectively. Since four MLP layers are paralleled to project the attribute values of the $n$ nodes, the time complexity is thus $\mathcal{O}(4nd\hat{d})$. For the quaternion graph convolution, the feature quaternion in size $n\times (4\times\hat{d})$ will be processed by $l$ stacked quaternion graph encoders. The parameters $\mathbf{W}^{Q}_l$ of each encoder are with the same scale as the feature quaternion. Hence, the time complexity of quaternion graph convolution is $\mathcal{O}(l(n4\hat{d})^2)$. For the graph reconstruction, the inner product is conducted on the matrix $\mathbf{\Gamma}$ with size $n\times\hat{d}$, which consumes $\mathcal{O}(n^2\hat{d})$. 
Assume the training of HyReaL iterates $T$ times, the overall time complexity is $\mathcal{O}(T[4nd\hat{d}+l(n4\hat{d})^2+n^2\hat{d}])$. By omitting the small constants and the terms with lower magnitude, the final complexity is nearly $\mathcal{O}(T[nd\hat{d}+n^2\hat{d}^2])$.

\begin{algorithm}[t]
    \caption{HyReaL: Hyper-complex space Representation Learning.}
    \label{alg:algorithm}
    \textbf{Input}: Attributed graph $G=\{\mathbf{A},\mathbf X\}$; Cluster number $k$; Loss weights $\alpha$ and $\beta$. \\ 
    \textbf{Output}: $k$ non-overlapping sub-graphs $\{G_1, G_2, ..., G_k\}$.
    \begin{algorithmic}[1] 
    \STATE Convert the adjacency matrix $\mathbf A$ into symmetric normalized Laplacian matrix $\mathbf{\tilde A}$; 
    \REPEAT  
    \STATE Project $\mathbf{X}$ into four views $\mathbf{F}_{\triangleright}$ by Eq. (\ref{multi}) and form a feature quaternion $\mathbf{F}$ as shown in Eq. (\ref{cat});
    \STATE Encode $\mathbf{F}$ using quaternion graph encoders defined by Eqs.~(\ref{quaternion-parameter}) and~(\ref{encoder});
    \STATE Obtain the output embeddings $\mathbf{\Gamma}$ by the quaternion fusion operator defined in Eq. (\ref{avg});
    \STATE Reconstruct the adjacency matrix $\mathbf{\hat{A}}$ from $\mathbf{\Gamma}$ according to Eq. (\ref{re});
    \STATE Compute the value of objective function $\mathcal{L}$ 
    according to Eqs.~(\ref{loss}) -~(\ref{sc-process});
    \STATE Update learnable parameters $\mathbf{W}^{L}_{\triangleright}, \mathbf{B}^{L}_\triangleright$ and $\mathbf{W}^{Q}_l$.
    \UNTIL{maximum iterations reached}
    \STATE Perform spectral clustering to solve Eq.~(\ref{sc-process}) based on $\mathbf{\hat{A}}$ reconstructed from the final $\mathbf{\Gamma}$.
    \end{algorithmic}
\end{algorithm}
\begin{figure}[!t]
    \centering
    \includegraphics[width=1\linewidth]{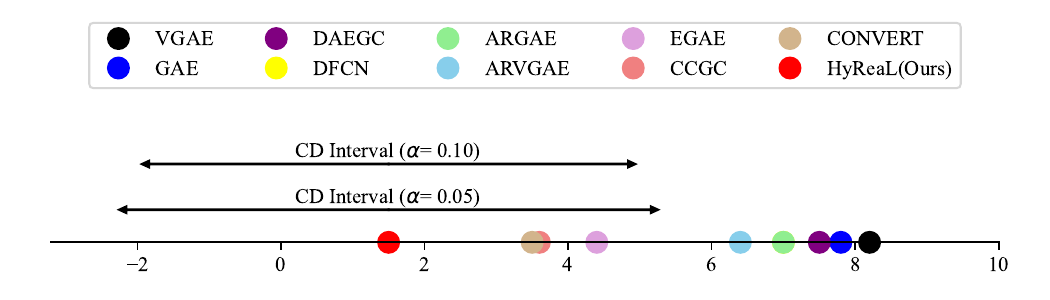}
    \caption{Visualization of Bonferroni-Dunn (BD) test at confidence intervals 90\% and 95\%.}
    \label{sig}
\end{figure}
\begin{figure*}[!t]
    \centering{\includegraphics[scale=0.44]{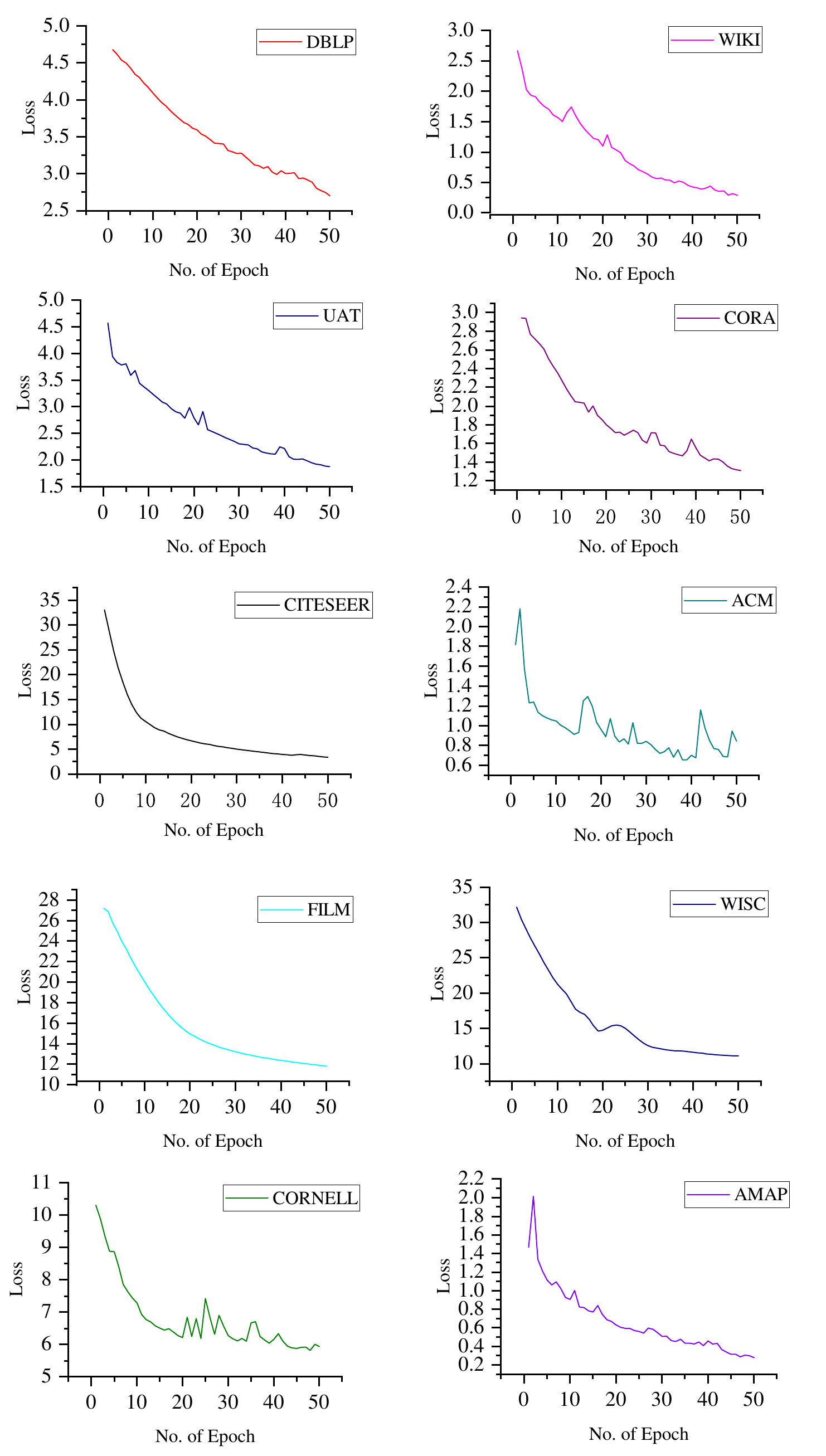}}
    \caption{Convergence curves of the HyReaL on ten datasets.}
    \label{loss-fig}
\end{figure*}

 \begin{figure*}[!t]
     \centering
         \includegraphics[width=0.9\textwidth]{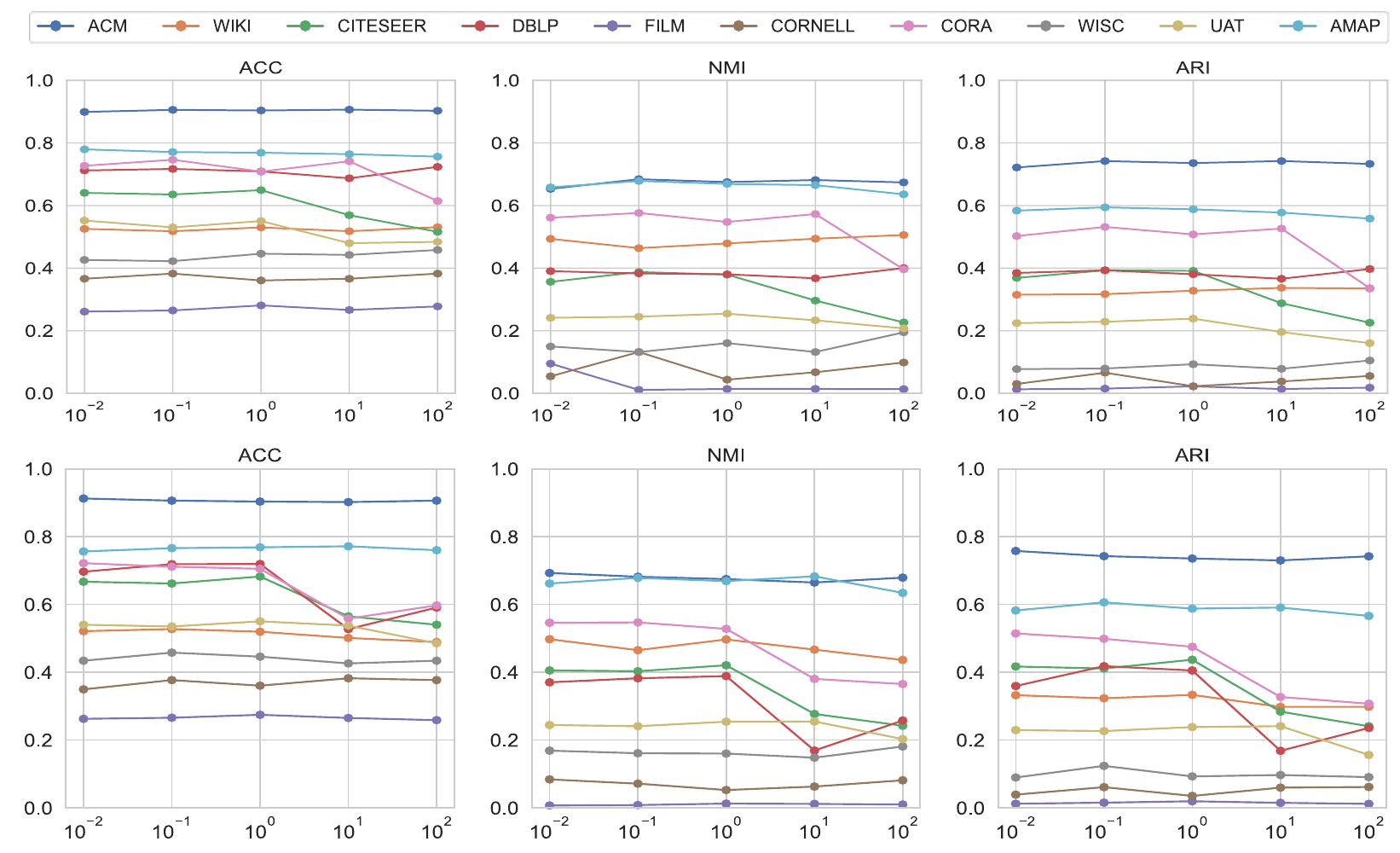}
     \caption{Sensitivity analysis of the trade-off parameters of the loss terms, i.e., $\alpha$ for $\mathcal{L}_{reg}$ (the upper row) and $\beta$ for $\mathcal{L}_{sc}$ (the lower row), on all the ten datasets (marked in lines with different colors). $x$-axes indicate the values of $\alpha$ and $\beta$.}
     \label{sensitivity}
 \end{figure*}
 \begin{figure*}[!t]
        \centering
    \includegraphics[width=1\linewidth]{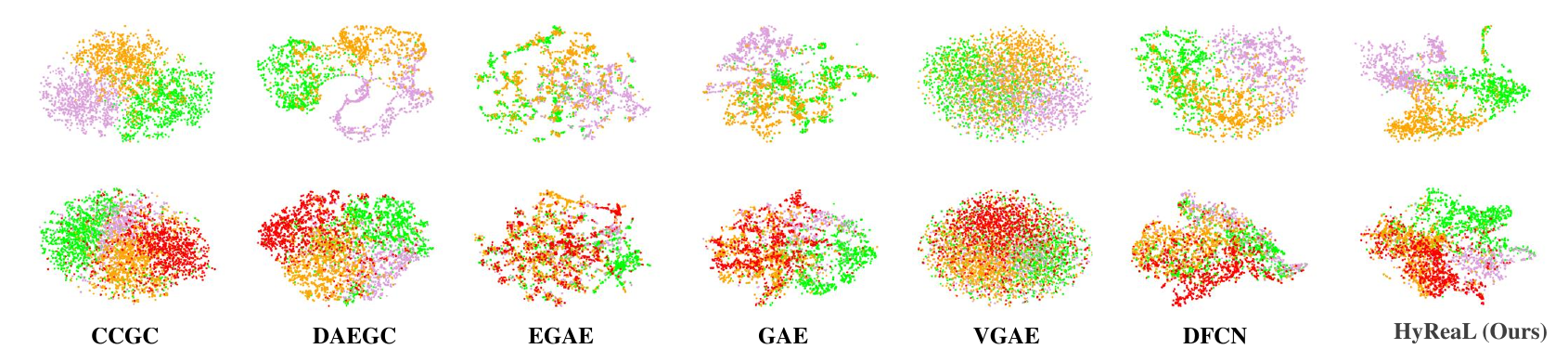}
    \caption{$t$-SNE visualization on ACM datasets. The first and second rows correspond to ACM and DBLP, respectively.}
    \label{append-visual}
\end{figure*}

\begin{figure*}[!t]
    \centering
    \includegraphics[width=1\linewidth]{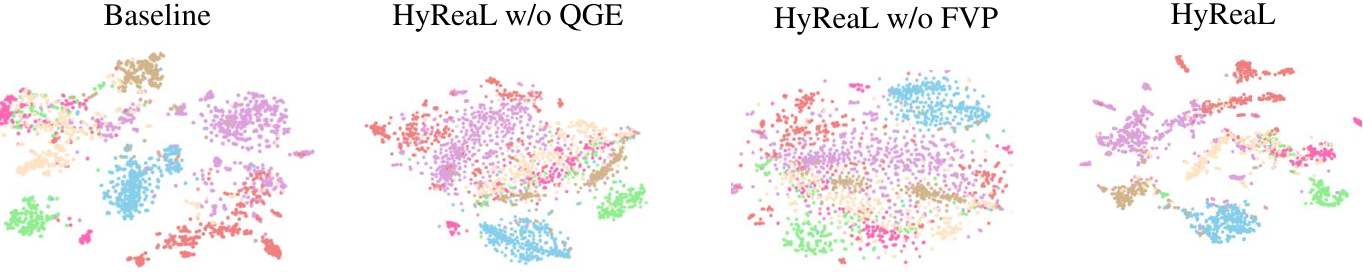}
    \caption{$t$-SNE visualization of the ablated variants of HyReaL on CORA dataset.}
    \label{fig:enter-label}
\end{figure*}

\section{Complementary Experimental Results}
\subsection{The complementary observation of External Metrics Evaluation}
In the following, three more key observations are provided: 

1) There are four performance groups of the compared methods:
Based on the average ranks in the ``AR'' row, there are four groups of methods with prominent AR gaps. The K-Means and Spectral Clustering (Spectral-C) with ARs around nine belong to the first group, as they are traditional methods without representation learning. The second groups would be the GAE, VGAE, ARGAE, ARVGAE, DFCN, and DAEGC with ARs within $[5.7,7.3]$. They are all based on the GAE and the latter two (i.e., DFCN and DAEGC) further incorporate clustering objectives for training. The third group consists of CONVERT, CCGC, and EGAE, all with an AR of around 4. The proposed HyReaL surely belongs to the fourth group with an AR close to one.

2) HyReaL vs. CCGC/CONVERT:
The proposed HyReaL achieves great performance improvements compared to the best-performing counterparts, which is usually the CCGC and CONVERT. Specifically, HyReaL outperforms the best-performing counterparts by 16.5\%, 121.2\%, 13.2\%, and 29.2\% on WIKI, DBLP, CORA, and WISC datasets, respectively, in terms of ARI. On most other datasets, our HyReaL also achieves considerable improvements of around 5\% in comparison with the rivals. Compared to our HyReaL, the CCGC and CONVERT methods adopt a contrastive learning paradigm by treating K-Means as the proxy task. Their data augmentation effectiveness relies on the selection of proper cluster number $k$, which is a non-trivial task, because the original $k$ provided by the dataset labels is not necessarily the `true' $k$. Accordingly, the performance of CCGC and CONVERT is relatively unstable on different datasets. 

3) HyReaL vs. EGAE:
EGAE adopts a relaxed K-Means to optimize the representation, which also requires a proper cluster number $k$, and thus achieves satisfactory clustering performance. Even though the training process of our HyReaL is not guided by  $k$, it still stably performs the best in most cases. The reason would be that even the `true' $k$ provided by the original dataset may still be unsuitable for the fusion of inconsistent attributes and graph topology. The use of a given $k$ can be viewed as introducing a strong hypothesis that may implicitly restrict the fitting ability of representation learning. By contrast, HyReaL adopts a relaxed clustering objective without restricting the node representations to be concentrated on $k$ potential clusters. As a result, HyReaL fosters a high DoF learning, and thus universal clustering-friendly representations can be obtained. 

\subsection{Complementary Separability and Universality Evaluation of Representations}
\begin{figure*}[!t]
    \centering
        \includegraphics[width=1\textwidth]{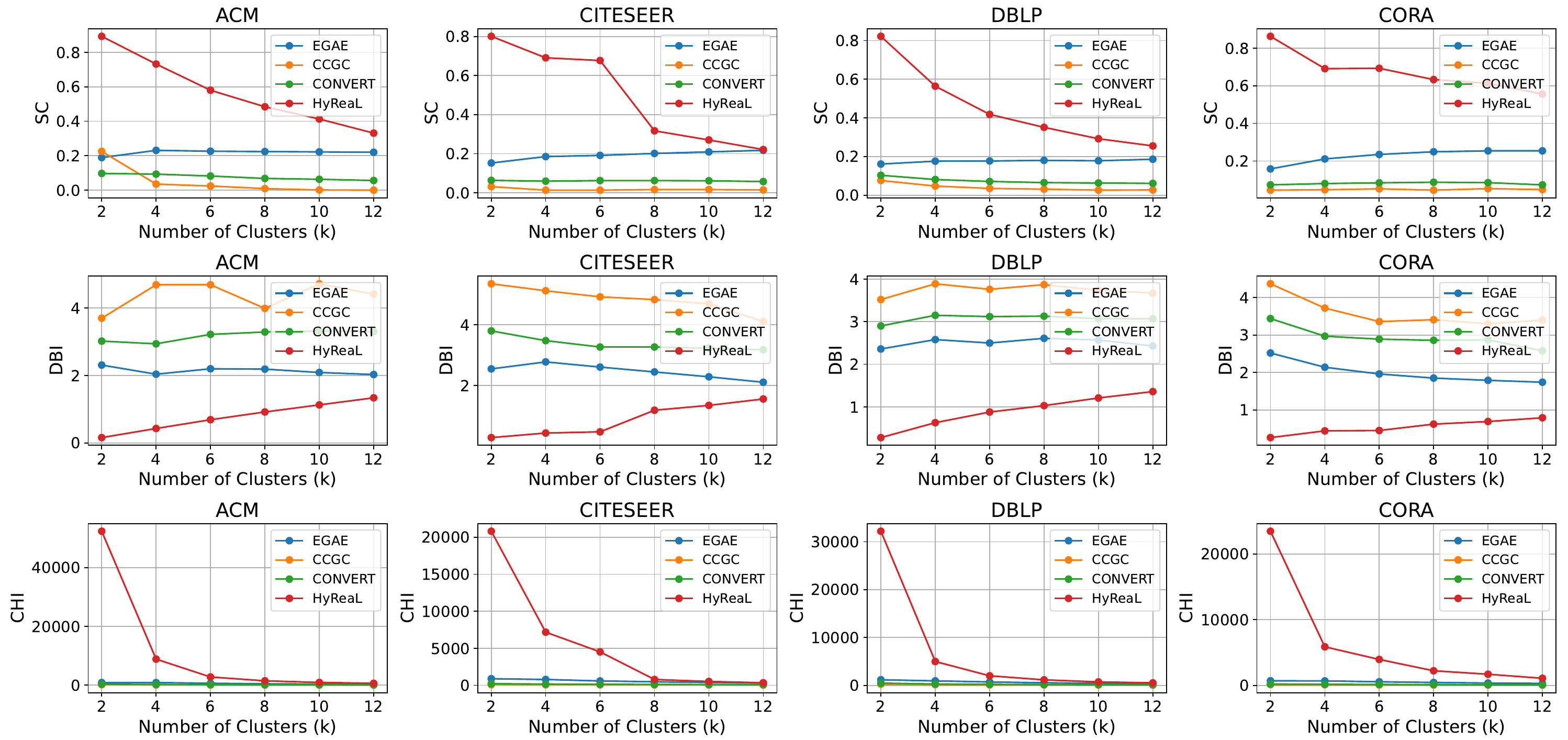}
    \caption{Clustering performance comparison using internal metrics under different $k$s. For the SC and CHI metrics, the higher the better. For the DBI metric, the lower the better.}
    \label{appendix-different-k-clustering}
\end{figure*}
The complementary experiments of separability and universality evaluation of representations, the plot is shown in Figure~\ref{appendix-different-k-clustering}.

\subsection{Complementary Effectiveness of QGE on Mitigating Over-smoothing}
\begin{figure*}
    \centering
        \includegraphics[width=1\textwidth]{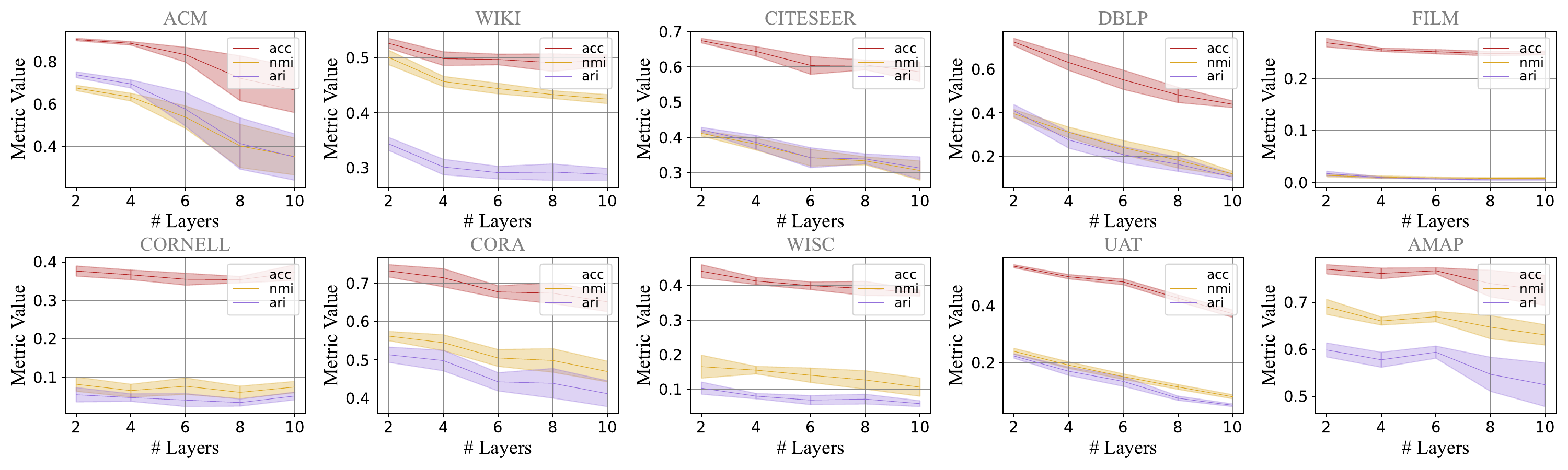}
    \caption{The OS effect analysis across ten datasets via QGE with varying layer numbers. When the number of layers increases, the OS effect significantly impacts clustering performance, highlighting the effectiveness of our method in alleviating OS.}
    \label{appendix-over-smoothing-ablation}
\end{figure*}
The complementary plots of effectiveness of QGE on mitigating over-smoothing, the plot is shown in Figure~\ref{appendix-over-smoothing-ablation}.

\subsection{Supplemental Ablation Studies}
Three observations are provided below:
1) HyReaL performs better than the Baseline in 29 out of 30 comparisons, clearly illustrating the adaptability of FVP and QGE modules in the learning.
2) HyReaL performs better than HyReaL w/o FVP in 27 out of 30 comparisons. This evidently indicates that FVP is a necessary pre-phase of QGE. HyReaL w/o FVP makes the four MLPs unlearnable, thus FVP degrades to a random projection of the input attributes, which surely loses the ability to provide suitable feature quaternions for QGE.
3) HyReaL outperforms HyReaL w/o QGE in 22 out of 30 comparisons. This indicates that QGE is effective in aggregating the node information and preventing the OD effect. Without the quaternion transformation in QGE, HyReaL cannot facilitate a high DoF representation learning of attributes to offset the OD effect. As a result, the embeddings of two very dissimilar but graph-adjacent nodes may be homogeneous, which hinders accurate clustering.
4) HyReaL performs better than HyReaL w/o $\beta$ in 23 out of 30 comparisons. This clearly claims that the cluster-oriented loss contributes to enhancing the representation ability. That is, the objective function provides more clustering-friendly optimization guidance in optimization.

\subsection{The Results of Wilcoxon Signed Rank Test on Comparative Experiments}
\label{app-subsec-wtest}
Table~\ref{wtest} is the Wilcoxon signed rank test of comparative experiments results.

\subsection{Bonferroni-Dunn Test of Comparison Experiment}
\label{app-subsec-BDtest}


In order to comprehensively demonstrate the superiority of our model compared to other methods, we conduct the Bonferroni-Dunn Test (BD test) \cite{DBLP:journals/jmlr/Demsar06} based on the average rank (i.e., the `AR' row) of the comparative experimental results in Table~1 of the main paper.

The Bonferroni-Dunn test is used to compare an algorithm with the remaining $k-1$ counterparts. It involves comparing the differences in average ranks of various methods with a certain threshold value called Critical Difference (CD). The CD is defined as:
\begin{equation}
    CD=q_{\lambda}{\sqrt{\frac{p(p+1)}{6N}}},
\end{equation}
where $q_{\lambda}$ is critical values for the BD test, $p$ is the number of compared methods, and $N$ is the number of dataset. If the rank difference between the two methods is higher than the CD, it indicates that the method with the higher average rank is statistically superior to the one with the lower average rank. Conversely, if the difference is lower than the CD, it suggests that there is no significant performance difference between the two methods.

Our BD test conduction procedures are as follows. 1) We obtain the ranks of the methods under all three validity metrics on all ten datasets. 2) The ranks under the three metrics are averaged to an overall rank of the corresponding method w.r.t. each certain dataset. 3) The average ranks on ten datasets are further averaged to an overall average rank of the methods, which are shown in Table~1 of the main paper.

According to \cite{DBLP:journals/jmlr/Demsar06}, we set the confidence intervals to 90\% and 95\%, and compute the CD by
\begin{equation}
 CD_{0.10}=3.4378, 
\end{equation}
and
\begin{equation}
    CD_{0.05}=3.7546,
\end{equation}
where the $q_{0.10}$ and $q_{0.05}$ of ten classifiers are 2.539 and 2.773 according to Table~5(b) in reference \cite{DBLP:journals/jmlr/Demsar06}, the number of datasets $N$ is 10, and the number of compared methods $p$ is 10. Overall, it can be observed that HyReaL performs significantly better than the seven methods, as shown in Figure \ref{sig}.

\subsection{Training Convergence Evaluation}
To demonstrate the convergence of our model, we show its convergence curves on all the ten benchmark datasets in Figure \ref{loss-fig}. 

The overall trend of the loss convergence curves indicates a steady decrease in loss, which suggests that the model can effectively learn from the training data. Although there are minor fluctuations in the loss curves on some datasets, the loss decreasing tends stable when approaching the pre-set 50 epoch of training. In summary, the training convergence evaluation illustrates that our model can be effectively trained for learning representation and clustering.

\subsection{Sensitivity Evaluation of Hyper-Parameters}
The sensitivity of HyReaL to the trade-off hyper-parameters $\alpha$ and $\beta$ is evaluated on the datasets as shown in Figure~\ref{sensitivity}. Note that when evaluating sensitivity to one parameter, another one is fixed at the corresponding settings in Appendix~\ref{model-settings}. From the results, it is not surprising that a too-large value of $\alpha$ or $\beta$ leads to generating objective biased representations such that HyReaL obtains undesired clustering performance. The results also confirmed that HyReaL is insensitive to $\alpha$ and $\beta$ in the value range around the parameter settings adopted for the aforementioned experiments.

\subsection{Visual Results}
The supplementary $t$-SNE visualization results of the representations learned by different methods on the ACM and DBLP datasets are shown in Fig~\ref{append-visual}. 

To intuitively compare the ablated versions of HyReaL, the representations learned by them and HyReaL are also compared using $t$-SNE on the CORA dataset in Figure~\ref{fig:enter-label}.

For all the visualization results in this section, the observations and conclusions are consistent with the corresponding results in the main paper, so we do not provide redundant discussions here.

\section{Discussion about Remark and Proof}
\label{app-sec-DoF}
\subsection{Detailed Remark of Learning Ability}
We provide a more detailed analysis of ``Remark 1'' in Section~\ref{section-HyReaL} of the submitted paper. The more detailed Remark 1 is given below.
\begin{remark}
\textbf{Degree of Freedom.}
According to Eq.~(\ref{quaternion-parameter}) in main paper, learnable parameters in our model, i.e., $\mathbf{W}^{Q}_{\mathbb{H}}=\{\mathbf{W}^{Q}_r,\mathbf{W}^{Q}_x,\mathbf{W}^{Q}_y,\mathbf{W}^{Q}_z\}$, yields 16 pairs of feature interaction. In contrast, realizing the same scale interaction in real-value space requires 4 times of parameters. This illustrates the learning efficiency of the proposed model. Detailed analysis is given below. 

Given model input 
\begin{equation}
\mathbf{F}=\{\mathbf{F}_r,\mathbf{F}_x,\mathbf{F}_y,\mathbf{F}_z\},
\end{equation}
where $\mathbf{F}\in\mathbb{H}^{n\times (4\times \hat{d})}$, $\hat{d}$ indicates the dimension of input. Then, we define the learnable parameters of quaternion representation as $\mathbf{W}^{Q}_{\mathbb{H}}\in \mathbb{H}^{(4\times\hat{d}) \times (4\times \tilde{d})}$, which contains four part of parameters $\{\mathbf{W}^{Q}_r,\mathbf{W}^{Q}_x,\mathbf{W}^{Q}_y,\mathbf{W}^{Q}_z\}$, where $\tilde d$ is the output dimension, and $\mathbf{W}^{Q}_i\in \mathbb{H}^{\hat{d}\times\Tilde{d}}$ with $i\in\{r,x,y,z\}$. 

According to the Hamilton product in the quaternion system, the learnable parameters let the features in $\mathbf{F}$ interact by
\begin{equation}
    \begin{aligned}
\mathbf{F}^{Q}
& =\mathbf{F} \otimes \mathbf{W}^{Q}\\
& =\mathbf{W}^{Q}_r \mathbf{F}_r  -\mathbf {W}^{Q}_x\mathbf{F}_x  -\mathbf{W}^{Q}_y\mathbf{F}_y-\mathbf{W}^{Q}_z \mathbf{F}_z \\
& +\mathbf{W}^{Q}_x\mathbf{F}_r  +\mathbf{W}^{Q}_r\mathbf{F}_x  -\mathbf {W}^{Q}_z\mathbf{F}_y+\mathbf{W}^{Q}_y\mathbf{F}_z \\
& +\mathbf{W}^{Q}_y\mathbf{F}_r  +\mathbf{W}^{Q}_z\mathbf{F}_x +\mathbf{W}^{Q}_r\mathbf{F}_y-\mathbf{W}^{Q}_x\mathbf{F}_z  \\
& +\mathbf{W}^{Q}_z\mathbf{F}_r  -\mathbf{W}^{Q}_y\mathbf{F}_x +\mathbf{W}^{Q}_x\mathbf{F}_y+\mathbf{W}^{Q}_r\mathbf{F}_z  
\end{aligned},
\end{equation}
where $\mathbf{F}^Q\in\mathbb{H}^{n\times (4\times\tilde{d})}$. It is intuitive that such an operation yields learning with a 16-Degree of Freedom (DoF). 

In the following, we design a real-value model with the same DoF, and observe how many parameters are required for comparison. For intuitive comparison, we define the parameters of the real-value model in a similar form as that of the quaternion model, i.e., $\mathbf{W}^{R}_i\in\mathbb{R}^{4\times( \hat{d}\times \tilde{d}) }$ with $i\in\{r,x,y,z\}$. The superscript $R$ indicates that these are the parameters of the real-value model. Accordingly, all the features in $\mathbf{F}$ interact through the parameters by
\begin{equation}
     \mathbf{F}^{R} =[\mathbf{F}_r \mathbf{W}^{R}_r,\mathbf{F}_x \mathbf{W}^{R}_x,\mathbf{F}_y \mathbf{W}^{R}_y,\mathbf{F}_z \mathbf{W}^{R}_z],
\end{equation}
where $\mathbf{F}^R\in\mathbb{R}^{n\times(4\times\Tilde{d})}$ is the output matrix, and $\mathbf{W}^{R}_r,\mathbf{W}^{R}_x,\mathbf{W}^{R}_y,\mathbf{W}^{R}_z$ can be written as
\begin{equation}
\begin{aligned}
    &\mathbf{W}^{R}_r=[\mathbf{W}^{R}_1,\mathbf{W}^{R}_2,\mathbf{W}^{R}_3,\mathbf{W}^{R}_4]\\
    &\mathbf{W}^{R}_x=[\mathbf{W}^{R}_5,\mathbf{W}^{R}_6,\mathbf{W}^{R}_7,\mathbf{W}^{R}_8]\\
    &\mathbf{W}^{R}_y=[\mathbf{W}^{R}_9,\mathbf{W}^{R}_{10},\mathbf{W}^{R}_{11},\mathbf{W}^{R}_{12}]\\
    &\mathbf{W}^{R}_z=[\mathbf{W}^{R}_{13},\mathbf{W}^{R}_{14},\mathbf{W}^{R}_{15},\mathbf{W}^{R}_{16}].  
\end{aligned}
\end{equation}
Accordingly, the output feature $\mathbf{F}^R$ is written as
\begin{equation}
\begin{aligned}
\mathbf{F}^{R} =\left[
\begin{matrix}
\mathbf{W}^{R}_1 \mathbf{F}_r  +\mathbf {W}^{R}_2\mathbf{F}_r  +\mathbf{W}^{R}_3\mathbf{F}_r+\mathbf{W}^{R}_4 \mathbf{F}_r \\
\mathbf{W}^{R}_5\mathbf{F}_x  +\mathbf{W}^{R}_6\mathbf{F}_x  +\mathbf {W}^{R}_7\mathbf{F}_x+\mathbf{W}^{R}_8\mathbf{F}_x \\
\mathbf{W}^{R}_9\mathbf{F}_y  +\mathbf{W}^{R}_{10}\mathbf{F}_y +\mathbf{W}^{R}_{11}\mathbf{F}_y+\mathbf{W}^{R}_{12}\mathbf{F}_y  \\
\mathbf{W}^{R}_{13}\mathbf{F}_z  +\mathbf{W}^{R}_{14}\mathbf{F}_z +\mathbf{W}^{R}_{15}\mathbf{F}_z+\mathbf{W}^{R}_{16}\mathbf{F}_z  
\end{matrix}
\right]^\top_{.}
\end{aligned}
\end{equation}

Obviously, there are 16 parameter matrices that are of the same size of $\mathbf{W}^{R}_i\in\mathbb{R}^{4\times( \hat{d}\times \tilde{d}) }$, and it can be concluded that if a real-value model is adopted to realize the same DoF as that of the quaternion-value model, a four-time model parameters scale will be involved. In other words, with the same number of parameters, the quaternion-value model can achieve a higher DoF than real-value models for more informative representation and cluster learning. 
\end{remark}

\subsection{Proof of Degree of Freedom}\label{appendix:DoF}
According to above intuitive discussions, the learnable parameters $\mathbf{W}^Q=\{\mathbf{W}^Q_r,\mathbf{W}^Q_x,\mathbf{W}^Q_y,\mathbf{W}^Q_z\}$ can generate 16 features learning pairs. In the same DoF representation, the number of real-value parameters is four times of the quaternion representation. Thus, with the same number of parameters, the quaternion representation model can achieve a four times DoF comparing to the conventional real-value representation learning, and thus helps to more thoroughly explore features informative and benefit cluster learning.

Rigorous proof of the `4$\times$DoF' is provided below. 
We follow the parameter initialization method in \cite{DOFproof} and initial our quaternion graph encoders.
In order to simplify the mathematical expression and analyses, especially when dealing with operations involving complex numbers or quaternions, we introduce the polar coordinate form to represent the weight of quaternion.
A generated quaternion weight $w$ from a weight matrix $\mathbf{W}^Q$ has a polar form defined as:
\begin{equation}
    w=|w|e^{q_{img}^{\triangleleft}\theta}=|w|(\cos(\theta)+q_{img}^{\triangleleft}\sin(\theta)), 
\end{equation}
with $q_{img}^{\triangleleft}=0+x \mathbf{i}+y \mathbf{j}+z \mathbf{k}$ a purely imaginary and normalized quaternion, and the angle $\theta$ is randomly generated in the interval $[-\pi,\pi]$. The imaginary components $x\mathbf{i}$, $y\mathbf{j}$, and $z\mathbf{k}$ are sampled from an uniform distribution in $[0,1]$ to obtian $q_{imag}$. The parameter $\varphi$ is a random number generated with respect to well-known initialization criterions (such as Glorot \cite{init} or He algorithms \cite{he-iccv}). Therefore, $w$ can be computed following:
\begin{equation}
    \begin{gathered}
w_{\mathbf{r}}=\varphi \cos (\theta), \\
w_{\mathbf{i}}=\varphi q_{img \mathbf{i}}^{\triangleleft} \sin (\theta), \\
w_{\mathbf{j}}=\varphi q_{img \mathbf{j}}^{\triangleleft} \sin (\theta), \\
w_{\mathbf{k}}=\varphi q_{img \mathbf{k}}^{\triangleleft} \sin (\theta) .
\end{gathered}
\end{equation}

However, $\varphi$ represents a randomly generated variable with respect to the variance of the quaternion weight and the selected initialization criterion. The initialization process follows \cite{init} or \cite{he-iccv} to derive the variance of the quaternion-valued weight parameters. 
Indeed, the variance of $\mathbf{W^Q}$ has to be investigated:

\begin{equation}
    \operatorname{Var}(\mathbf{W^Q})=\mathbb{E}\left(|\mathbf{W^Q}|^{2}\right)-[\mathbb{E}(|\mathbf{W^Q}|)]^{2},
\end{equation}
\textbf{$[\mathbb{E}(|\mathbf{W^Q}|)]^{2}$ equals to 0 since the weight distribution is symmetric around 0.} Nonetheless, the value of $\operatorname{Var}(\mathbf{W^Q})=\mathbb{E}\left(|\mathbf{W^Q}|^{2}\right)$ is not trivial in the case of quaternion-valued matrices. Indeed, $\mathbf{W^Q}$ follows a Chi-distribution with four degrees of freedom (DoFs). Chi-distribution is often used to describe the distribution mode of the modulus length or Euclidean distance of multi-dimensional vectors. Thus, $\mathbb{E}\left(|\mathbf{W^Q}|^{2}\right)$ is expressed and computed as follows:
\begin{equation}
    \mathbb{E}\left(|\mathbf{W^Q}|^{2}\right)=\int_{0}^{\infty} x^{2} f(x) \mathrm{d} x,
\end{equation}
with $f(x)$ is the probability density function with four DoFs. A four-dimensional vector $X=$ $\{A, B, C, D\}$ is considered to evaluate the density function $f(x)$. $X$ has components that are normally distributed, centered at zero, and independent. Then, $A, B, C$ and $D$ have density functions:
\begin{equation}
    f_{A}(x ; \sigma)=f_{B}(x ; \sigma)=f_{C}(x ; \sigma)=f_{D}(x ; \sigma)=\frac{e^{-x^{2} / 2 \sigma^{2}}}{\sqrt{2 \pi \sigma^{2}}} .
\end{equation}

The four-dimensional vector $X$ has a length $L$ defined as $L=\sqrt{A^{2}+B^{2}+C^{2}+D^{2}}$ with a cumulative distribution function $F_{L}(x ; \sigma)$ in the 4-sphere (n-sphere with $n=4$) $S_{x}$:
\begin{equation}
    F_{L}(x ; \sigma)=\iiiint_{S_{x}} f_{A}(x ; \sigma) f_{B}(x ; \sigma) f_{C}(x ; \sigma) f_{D}(x ; \sigma) \mathrm{d} S_{x},
    \label{30F}
\end{equation}
where $S_{x}=\left\{(a, b, c, d): \sqrt{a^{2}+b^{2}+c^{2}+d^{2}}<x\right\}$ and $\mathrm{d} S_{x}=\mathrm{d} a \mathrm{d} b \mathrm{d} c \mathrm{d} d$. The polar representations of the coordinates of $X$ in a 4-dimensional space are defined to compute $\mathrm{d} S_{x}$ :
\begin{equation}
    \begin{aligned}
& a=\rho \cos \theta \\
& b=\rho \sin \theta \cos \phi \\
& c=\rho \sin \theta \sin \phi \cos \psi \\
& d=\rho \sin \theta \sin \phi \sin \psi,
\end{aligned}
\end{equation}
where $\rho$ is the magnitude ( $\rho=\sqrt{a^{2}+b^{2}+c^{2}+d^{2}}$ ) and $\theta, \phi$, and $\psi$ are the phases with $0 \leq \theta \leq \pi$, $0 \leq \phi \leq \pi$ and $0 \leq \psi \leq 2 \pi$. Then, $\mathrm{d} S_{x}$ is evaluated with the Jacobian $J_{f}$ of $f$ defined as:
\begin{equation}
    \begin{aligned}
& J_{f}=\frac{\partial(a, b, c, d)}{\partial(\rho, \theta, \phi, \psi)}=\frac{\mathrm{d} a \mathrm{d} b \mathrm{d} c \mathrm{d} d}{\mathrm{d} \rho \mathrm{d} \theta \mathrm{d} \phi \mathrm{d} \psi}=
\left|
\begin{array}{llll}
\frac{\mathrm{d} a}{\mathrm{d} \rho} & \frac{\mathrm{d} a}{\mathrm{d} \theta} & \frac{\mathrm{d} a}{\mathrm{d} \phi} & \frac{\mathrm{d} a}{\mathrm{d} \psi} \\
\frac{\mathrm{d} b}{\mathrm{d} \rho} & \frac{\mathrm{d} b}{\mathrm{d} \theta} & \frac{\mathrm{d} b}{\mathrm{d} \phi} & \frac{\mathrm{d} b}{\mathrm{d} \psi} \\
\frac{\mathrm{d} c}{\mathrm{d} \rho} & \frac{\mathrm{d} c}{\mathrm{d} \theta} & \frac{\mathrm{d} c}{\mathrm{d} \phi} & \frac{\mathrm{d} c}{\mathrm{d} \psi} \\
\frac{\mathrm{d} d}{\mathrm{d} \rho} & \frac{\mathrm{d} d}{\mathrm{d} \theta} & \frac{\mathrm{d} d}{\mathrm{d} \phi} & \frac{\mathrm{d} d}{\mathrm{d} \psi}
\end{array}
\right| \\
& =\left|\begin{array}{cccc}
\cos \theta & -\rho \sin \theta & 0 & 0 \\
\sin \theta \cos \phi & \rho \sin \theta \cos \phi & -\rho \sin \theta \sin \phi & 0 \\
\sin \theta \sin \phi \cos \psi & \rho \cos \theta \sin \phi \cos \psi & \rho \sin \theta \cos \phi \cos \psi & -\rho \sin \theta \sin \phi \sin \psi \\
\sin \theta \sin \phi \sin \psi & \rho \cos \theta \sin \phi \sin \psi & \rho \sin \theta \cos \phi \sin \psi & \rho \sin \theta \sin \phi \cos \psi
\end{array}\right|.
\end{aligned}
\end{equation}
And,
\begin{equation}
    J_{f}=\rho^{3} \sin ^{2} \theta \sin \phi.
\end{equation}
Therefore, by the Jacobian $J_{f}$, we have the polar form:
\begin{equation}
    \mathrm{d} a \mathrm{d} b \mathrm{d} c \mathrm{d} d=\rho^{3} \sin^{2}\theta \sin \phi \mathrm{d} \rho \mathrm{d} \theta \mathrm{d} \phi \mathrm{d} \psi.
\end{equation}
Then, writing Eq. (\ref{30F}) in polar coordinates, we obtain:
\begin{equation}
    \begin{aligned}
F_{L}(x, \sigma) & =\left(\frac{1}{\sqrt{2 \pi \sigma^{2}}}\right)^{4} \iiint\int_{0}^{x} e^{-a^{2} / 2 \sigma^{2}} e^{-b^{2} / 2 \sigma^{2}} e^{-c^{2} / 2 \sigma^{2}} e^{-d^{2} / 2 \sigma^{2}} \mathrm{d} S_{x} \\
& =\frac{1}{4 \pi^{2} \sigma^{4}} \int_{0}^{2 \pi} \int_{0}^{\pi} \int_{0}^{\pi} \int_{0}^{x} e^{-\rho^{2} / 2 \sigma^{2}} \rho^{3} \sin ^{2} \theta \sin \phi \mathrm{d} \rho \mathrm{d} \theta \mathrm{d} \phi \mathrm{d} \psi \\
& =\frac{1}{4 \pi^{2} \sigma^{4}} \int_{0}^{2 \pi} \mathrm{d} \psi \int_{0}^{\pi} \sin \phi \mathrm{d} \phi \int_{0}^{\pi} \sin ^{2} \theta \mathrm{d} \theta \int_{0}^{x} \rho^{3} e^{-\rho^{2} / 2 \sigma^{2}} \mathrm{d} \rho \\
& =\frac{1}{4 \pi^{2} \sigma^{4}} 2 \pi 2\left[\frac{\theta}{2}-\frac{\sin 2 \theta}{4}\right]_{0}^{\pi} \int_{0}^{x} \rho^{3} e^{-\rho^{2} / 2 \sigma^{2}} \mathrm{d} \rho \\
& {=\frac{1}{4 \pi^{2} \sigma^{4}} 4 \pi \frac{\pi}{2} \int_{0}^{x} \rho^{3} e^{-\rho^{2} / 2 \sigma^{2}} \mathrm{d} \rho}_{.}
\end{aligned}
\end{equation}
Then,
\begin{equation}
    F_{L}(x, \sigma)=\frac{1}{2 \sigma^{4}} \int_{0}^{x} \rho^{3} e^{-\rho^{2} / 2 \sigma^{2}} \mathrm{d} \rho .
\end{equation}
The probability density function for $X$ is the derivative of its cumulative distribution function, which by the fundamental theorem of calculus is:
\begin{equation}
    \begin{aligned}
f_{L}(x, \sigma) & =\frac{\mathrm{d}}{\mathrm{d} x} F_{L}(x, \sigma) \\
& {=\frac{1}{2 \sigma^{4}} x^{3} e^{-x^{2} / 2 \sigma^{2}}}_{.}
\end{aligned}
\end{equation}
The expectation of the squared magnitude becomes:
\begin{equation}
    \begin{aligned}
\mathbb{E}\left(|\mathbf{W}^Q|^{2}\right) & =\int_{0}^{\infty} x^{2} f(x) \mathrm{d} x \\
& =\int_{0}^{\infty} x^{2} \frac{1}{2 \sigma^{4}} x^{3} e^{-x^{2} / 2 \sigma^{2}} \mathrm{d} x \\
& {=\frac{1}{2 \sigma^{4}} \int_{0}^{\infty} x^{5} e^{-x^{2} / 2 \sigma^{2}} \mathrm{d} x} _{.}
\end{aligned}
\end{equation}
With integration by parts we obtain:
\begin{equation}
    \begin{aligned}
\mathbb{E}\left(|\mathbf{W}^Q|^{2}\right) & =\frac{1}{2 \sigma^{4}}\left(-\left.x^{4} \sigma^{2} e^{-x^{2} / 2 \sigma^{2}}\right|_{0} ^{\infty}+\int_{0}^{\infty} \sigma^{2} 4 x^{3} e^{-x^{2} / 2 \sigma^{2}} \mathrm{d} x\right) \\
& =\frac{1}{2 \sigma^{2}}\left(-\left.x^{4} e^{-x^{2} / 2 \sigma^{2}}\right|_{0} ^{\infty}+\int_{0}^{\infty} 4 x^{3} e^{-x^{2} / 2 \sigma^{2}} \mathrm{d} x\right)_{.}
\end{aligned}
\label{35EW}
\end{equation}
The expectation $\mathbb{E}\left(|\mathbf{W}^Q|^{2}\right)$ is the sum of two terms. The first one:
\begin{equation}
    \begin{aligned}
-\left.x^{4} e^{-x^{2} / 2 \sigma^{2}}\right|_{0} ^{\infty} & =\lim _{x \rightarrow+\infty}-x^{4} e^{-x^{2} / 2 \sigma^{2}}-\lim _{x \rightarrow+0} x^{4} e^{-x^{2} / 2 \sigma^{2}} \\
& {=\lim _{x \rightarrow+\infty}-x^{4} e^{-x^{2} / 2 \sigma^{2}}}_{.}
\end{aligned}
\end{equation}
Based on the L'H$\mathrm{\hat{o}}$pital's rule, the undetermined limit becomes:
\begin{equation}
    \begin{aligned}
    \operatorname*{lim}_{x \rightarrow +\infty}-x^{4}e^{-x^{2}/2\sigma^{2}}
    &=-\operatorname*{lim}_{x \rightarrow +\infty}\frac{x^{4}}{e^{x^{2}/2\sigma^{2}}}\\
    &= \,\,...\\
    &=-\operatorname*{lim}_{x \rightarrow +\infty}\frac{24}{(1/\sigma^{2})(P(x)e^{x^{2}/2\sigma^{2}})}\\
    &=0.
    \end{aligned}
    \label{36lim}
\end{equation}
With $P(x)$ is polynomial and has a limit to ${\mathbf{+\infty}}.$ The second term is calculated in a same way (integration by parts) and $\mathbb{E}(|\mathbf{W}^Q|^{2})$ becomes from Eq. (\ref{35EW}):
\begin{equation}
    \begin{aligned}
    \mathbb{E}(|\mathbf{W}^Q|)^{2}&=\frac{1}{2\sigma^{2}}\int_{0}^{\infty}4x^{3}e^{-x^{2}/2\sigma^{2}}\,\mathrm{d}x \\
    &=\displaystyle\frac{2}{\sigma^{2}}\left(x^{2}\sigma^{2}e^{-x^{2}/2\sigma^{2}}\Big|_{0}^{\infty}+\int_{0}^{\infty}\sigma^{2}2x e^{-x^{2}/2\sigma^{2}}\,\mathrm{d}x\right)_{.}
    \end{aligned}
\end{equation}
The limit of first term is equals to 0 with the same method than in Eq. (\ref{36lim}). Therefore, the expectation is:
\begin{equation}
\begin{aligned}
    \mathbb{E}(|\mathbf{W}^Q|^{2})
    &=4\left(\int_{0}^{\infty}x e^{-x^{2}/2\sigma^{2}}\,\mathrm{d}x\right)\\ 
    &{=4\sigma^{2}}{.}
\end{aligned}
\end{equation}
And finally, the variance is:
\begin{equation}
    Var(|\mathbf{W}^Q|)=4\sigma^{2}. 
\end{equation}

By analyzing their probability distribution, we demonstrate that the degrees of freedom for quaternion weights in encoders are four times higher than those of conventional graph encoder weights, providing enhanced representational capacity.
\end{document}